# Tabular Embeddings for Tables with Bi-Dimensional Hierarchical Metadata and Nesting

Gyanendra Shrestha
Florida State University
Tallahassee, Florida, USA

Chutain Jiang
Florida State University
Tallahassee, Florida, USA

Sai Akula
Florida State University
Tallahassee, Florida, USA

Vivek Yannam
Florida State University
Tallahassee, Florida, USA

Anna Pyayt
University of South Florida
Tampa, Florida, USA

Michael Gubanov
Florida State University
Tallahassee, Florida, USA

## ABSTRACT

Embeddings serve as condensed vector representations for real-world entities, finding applications in Natural Language Processing (NLP), Computer Vision, and Data Management across diverse downstream tasks. Here, we introduce novel specialized embeddings optimized, and explicitly tailored to encode the intricacies of complex 2-D context in tables, featuring horizontal, vertical hierarchical metadata, and nesting. To accomplish that we define the Bi-dimensional tabular coordinates, separate horizontal, vertical metadata and data contexts by introducing a new visibility matrix, encode units and nesting through the embeddings specifically optimized for mimicking intricacies of such complex structured data. Through evaluation on 5 large-scale structured datasets and 3 popular downstream tasks, we observed that our solution outperforms the state-of-the-art models with the significant MAP delta of up to 0.28. GPT-4 LLM+RAG slightly outperforms us with MRR delta of up to 0.1, while we outperform it with the MAP delta of up to 0.42.

## 1 INTRODUCTION

Embeddings are dense numerical representations of real-world objects, expressed as vectors. In NLP and Information Retrieval (IR), embedding vectors commonly correspond to terms in text, and the corresponding vector space is expected to quantify the semantic similarity between them. While not the first, LLMs such as GPT-4 [54], Llama2 [74], FLAN-T5 [17] and others also heavily depend on embeddings. These trained models, sometimes also referred to as Generative AI or GenAI, store millions of such vectors, which are used to generate the response to the user's question. More recently, GenAI models were also trained for images – e.g., DALL – E2 and videos [39]. Other methods adapt NLP embeddings to obtain embeddings for relational tables [21]. Even though some of these approaches such as transformers [5, 22, 87] attend to every token in all sections of a table, including metadata and data, hence implicitly encode its 2D context, they are not explicitly optimized for complex structured data. To mitigate this limitation, there were a series of efforts [4, 12, 33, 36, 87] to construct more accurate embeddings capturing the intricacies of relational data. These include adding specialized embedding layers or an attention mechanism and pre-training the models on tasks such as table cells or segments recovery [21, 33, 36, 80], thereby making them aware of the tabular structure. The majority of these efforts are devoted only to relational and spreadsheet tables (0.9% and 22% of all tables in the Common Web Crawl [46]) [16]. They overlook the other widely used type of tables that we refer to as "non-relational". Unlike relational, they can exhibit not only single-header horizontal, but also multi-level hierarchical vertical, horizontal metadata, as well as nested tables [40]. This creates a gap in understanding these widely used tables in practice. Hence, it is important to take steps to bridge it by enabling machine table understanding for such tables. Several recent attempts [22, 23, 29, 44, 63] try to identify hierarchies and classify cells in such tables. However, these approaches are supervised, and labeling large amounts of such structured data is labor intensive, especially for large-scale datasets. Other recent approaches [14, 21, 33, 37, 48, 75, 84, 87], despite being unsupervised, are optimized for *relational* or primitive *non-relational* tables (e.g., *matrix* tables having just *singular* non-hierarchical metadata without the rest of their more complex features).

In 1st Normal Form [18] a relational table has a set of labeled homogeneous columns, which is not the case for the majority of tables in the real world, especially in medical, financial, and government tables. For example, Figure 1 illustrates such table detailing treatment efficacy from colorectal cancer. The lowest right cell has both horizontal (*Efficacy End Point → Other Efficacy*) and vertical (*Patient Cohort → Failing under Fluoropyrimidine and Irinotecan*) *hierarchical* metadata. Some cells have separate nested tables, all having values in different units, sometimes numerical ranges or gaussians [77].

Inspired by these major differences compared to more primitive tables, we introduce TabBiN — a novel self-supervised, transformer-based architecture to train fine-grained structurally-aware embeddings, *optimized* for tables with Bi-dimensional hierarchical metadata and Nesting. During pre-training the data cells from such tables are encoded using our novel *Bi-dimensional hierarchical* coordinates calculated based on their hierarchical and spatial in-table location. Different from the uni-tree structure [67, 80], TabBiN supports both explicit and implicit coordinate encodings, including those for nested tables with their own separate metadata, such as in Figure 1. To enable vertical, horizontal metadata, and data to efficiently aggregate their local neighboring 2D contexts, we propose a *metadata-aware attention mechanism* that is different from the regular transformer practices of bottom-up attention [61], and constituent attention [79] in NLP domain. We also adopt the Masked Language Model (MLM) pre-training objective from BERT [22] and Cell-level Cloze (CLC) to learn the representations of tokens and cells across a large volume of tables. We make the following contributions in this paper:

- For non-relational tables, not in 1st Normal Form, exhibiting hierarchical vertical, horizontal metadata, and nesting [40], we propose tabular Bi-dimensional hierarchical coordinates (see Figure 1). Using these coordinates, we devise a self-supervised, transformer-based, metadata-aware attention mechanism and pre-training method, key in creating our novel structurally-aware composite embeddings, optimized for such non-relational tables. During pre-training or fine-tuning TabBiN learns these embedding vectors representing cells, tuples, columns, horizontal, vertical metadata, or the entire table from large-scale corpora in self-supervised manner.

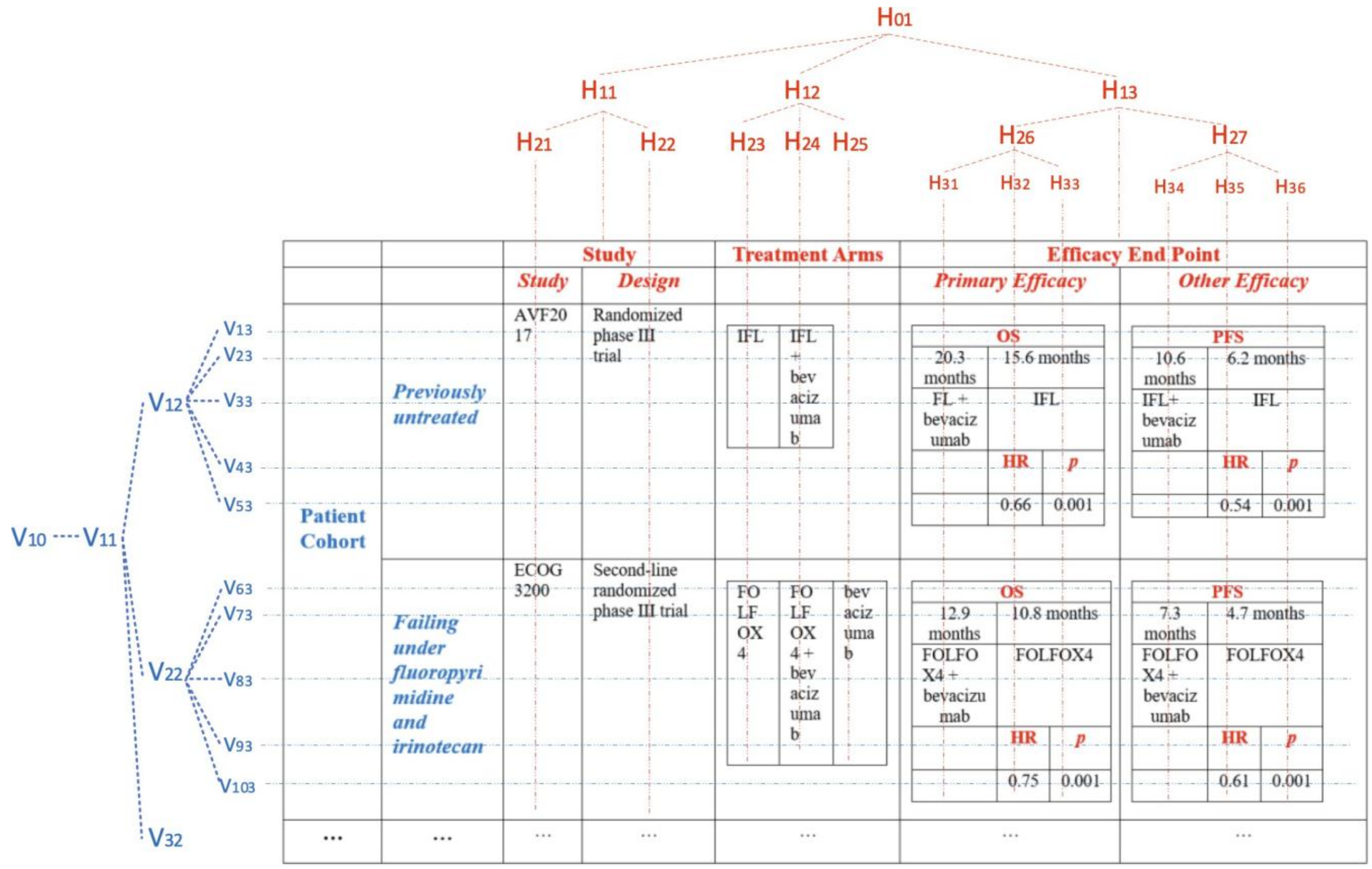

**Figure 1: Bi-dimensional Coordinates for a non-1st Normal Form Table with Hierarchical Metadata and Nesting.**

- To better incorporate semantics and intricacies of such complex structured data we introduce new abstractions, optimized for and explicitly encoding nested tables, entity types, units and ranges (for numerical data) used in the embedding layer of our architecture (see Figure 2).
- We fine-tune our embeddings on 5 large-scale structured datasets, evaluate and demonstrate that TabBiN outperforms or matches the state-of-the-art (SOTA) in most cases on 3 popular downstream tasks at scale. GPT-4+RAG slightly outperforms us with MRR delta of 0.1, while we outperform it with MAP delta up to 0.42.

Our *downstream tasks* are *column clustering*, *table clustering*, and *entity clustering/matching*. The clusters are inherently useful, for example, to find tables similar to a given table (based on the cosine similarity calculated using our embeddings) and can be used to aid table search [9–11, 19, 25, 27, 69], data fusion [10, 11, 57, 59, 69], taxonomy generation, and other tasks [71].

The remainder of the paper is structured as follows. Section 2 describes comprehensive definition of the problem, the datasets, Bi-dimensional coordinates. Section 3 details the TabBiN model, the structure of its composite embedding layer, and the pre-training methodology. Section 4 describes our experimental evaluation on 5 large-scale structured datasets having a variety of both relational and non-relational tables on 3 popular downstream tasks. We review related work in Section 5 and conclude in Section 6.

# 2 PRELIMINARIES

## 2.1 Definitions

The following definitions have been taken verbatim from our group's publication [40].

*Relational tables* [18], have the following properties: values are atomic, each column has values of the same type, each column has unique name (i.e., attribute name). The set of all attribute names is called table schema or metadata.

*Def* 1: Metadata is a set of *attributes* of a table. Metadata can be stored in a row - e.g., rows №1-2 in Figure 1, or in a column - e.g., columns №1-2 in Figure 1.

*Def* 2: Cell is a data value (i.e., can be a number, string, etc.) found at the intersection of a row and a column in a table. A relational table has C∗R cells total, where C is the number of columns and R is the number of rows.

*Def* 3: A table with hierarchical metadata is a table that, similar to a relational table, has metadata (i.e., attributes), but unlike a relational table it may be found not only in a row, but also in a column. It may also take several rows or columns. Such rows with metadata are called *horizontal metadata (HMD)*. On the other hand, such columns with metadata are called *vertical metadata (VMD)* [40].

**Table 1: Sample non-1NF Table with Nesting.**

<table>
<tr><th></th><th>Tumor Location</th><th>State</th><th>Primary Efficacy</th></tr>
<tr><td>Cancer</td><td>Colon</td><td>Florida</td><td><table><tr><th colspan="2">OS</th></tr><tr><td>20.3 months</td><td>15 months</td></tr><tr><td>bevacizumab</td><td>IFL</td></tr></table></td></tr>
</table>

We refer to tables not in 1$^{st}$ Normal Form (NF) [18], with Bi-dimensional hierarchical metadata and Nested tables inside cells as non-relational or BiN tables (see Table 1). Please refer to [40] for more formal definitions. Such tables often contain summary/aggregate data but are not limited to it [77].

A table in our work is represented as $T = [C, H, V, D]$, where $C$ is the table caption, which is a short text description summarizing what the table is about, $H = [c_1, c_2, c_3, ..., c_m]$ are $m$ columns in HMD, $V = [r_1, r_2, r_3, ..., r_n]$ are $n$ rows in VMD, $D = \{d_{ij} \mid 1 \leq i \leq n, 1 \leq j \leq m\}$ represent data cells, and $d_{ij}$ is the data cell in the $i^{th}$ row and $j^{th}$ column that has several tokens (texts or numbers). Given a table $T$, our embedding layer aims to learn in an unsupervised manner a structure-aware contextualized vector representation for each token in table cells to capture intricacies of 2D context within $T$. Specifically, we introduced new additional components in the embedding layer, encoding cell coordinates, nested tables, entity types, units, and ranges for better understanding of non-relational tabular data. These components are absent in existing transformer architectures for tabular data.

We now define the three table-related downstream tasks that we address in this paper.

**Column Clustering (CC).** The problem of pairwise column/attribute matching is well-known in schema matching [3, 16, 28], because these correspondences play a key role in identifying how to fuse two tables (i.e., which columns can be merged). This task involves the identification of similar $c_j \in H$ between two tables.

**Table Clustering (TC)**. TC is a task of grouping tables by topic (e.g., all Songs tables). This is a key task supporting table search, data fusion, where information from multiple tables on the same topic, originating from various sources, has to be integrated to provide a unified, comprehensive view [3, 16, 28].

**Entity Clustering/Matching (ECM).** Entity matching [42, 49] plays a crucial role in data fusion tasks by facilitating the identification and linkage of entities across disparate datasets. It establishes connections between entities from different sources, enabling a more comprehensive and accurate view of the data.

## 2.2 Datasets

To ensure we have a wide variety of tables we use 5 large-scale structured datasets. These datasets include both relational and non-relational tables.

- *Webtables* [46]: we took a sample of 20,000 tables in English including both relational and complex non-relational tables. On average, the tables have 14.45 rows and 5.2 columns. The most frequent topics covered in these tables include magazines, cities, universities, soccer clubs, regions, baseball players, and music genres. The cell values contain strings and numbers with and without units and ranges.
- *CovidKG* is a subset of CORD-19 [56], a public COVID-19 research dataset. We took a sample of 20,000 tables, related to COVID-19 and its vaccination, such as Moderna, Covaxin, Alpha variant, and Gamma variant. The table columns exhibit both VMD and HMD. The cell values contain strings, numbers with and without units, ranges, Gaussians, and nested tables.
- *CancerKG* dataset has 44,523 tables, extracted from all recent medical publications (up to 12/2023) on colorectal cancer, obtained via PubMed.com. The tables have 227,279 columns total, exhibiting both hierarchical VMD and HMD. The cell values contain strings, numbers with and without units, ranges, Gaussians, and nested tables.
- The 2010 *Statistical Abstract of the United States (SAUS)* comprises 1,320 tables [30, 80], which can be downloaded from the U.S. Census Bureau. The tables have 52.5 rows and 17.7 columns on average. It covers a variety of topics, including finance, business, crime, agriculture, and health care.
- The *CIUS* dataset [30, 80] is from the *Crime In the US (CIUS)* database and consists of 489 tables. The tables have 68.4 rows and 12.7 columns on average.

The non-relational tables that we defined in this paper are prevalent in our two datasets, CancerKG and CovidKG, constituting over 40% of each dataset. Additionally, approximately 10% of these complex tables exhibit nested structures in both datasets [77]. On average, the complex tables in our datasets consist of approximately 12 rows and 10 columns.

## 2.3 Bi-dimensional Coordinates

Figure 1 illustrates the Bi-dimensional coordinates that we introduce for non-relational tables, not in 1st Normal Form with hierarchical vertical and horizontal metadata, with nesting, defined in [40]. Our coordinates correspond to the cell location and the path through the metadata hierarchy to the cell. There are two coordinate-trees – *horizontal* and *vertical* (on the left and top of Figure 1). Both coordinate values correspond to the paths from the root nodes of the trees to the cell. For example, the coordinates of the table, nested in the upper right cell in Figure 1 (*Efficacy End Point → Other Efficacy*; *Patient Cohort → Previously Untreated*) are (<2,7>;<1,3>). In turn, the coordinates of the second horizontal metadata label (HR) in the nested table are (<3,5>;<4,3>). Notice that our bi-dimensional coordinates also apply to *relational* tables, whereby they reduce to the regular Cartesian coordinates. Tables in our corpora as well as in large-scale structured datasets in general usually come with unlabeled or noisy metadata. We designed and trained our own binary metadata classifiers based on Deep-learning *bi-GRU* and *CNN* architectures specifically for highly accurate labeling of multi-layer metadata - both horizontal and vertical [40]. One can also use other existing techniques for labeling metadata [50, 63].

# 3 TABBIN MODEL

Here we shed some light and provide more details on our Transformer-based self-supervised architecture with metadata-aware structural attention [5, 22] that we created for non-1st normal form tables with nesting and hierarchical metadata. Each encoder block in the Transformer is composed of a multi-head-self-attention layer and fully connected layer [5]. The configuration of our N-layer Transformer encoder model is aligned with $\text{BERT}_{BASE}$ [22]. However, we changed the standard BERT multi-head attention form $Q, K, V \in \mathbb{R}^{H \times H}$ [5], with our metadata-aware mask attention as follows:

$$TabBiN\ Attention(Q, K, V) = Attention(Q, K, V) \cdot M \tag{1}$$

where $H = 768$ is the hidden size of $\text{BERT}_{BASE}$ and $M \in \mathbb{R}^{n \times n}$ is the visibility matrix ($n$ in $M$ is the input sequence length), which we describe below in Section 3.2.

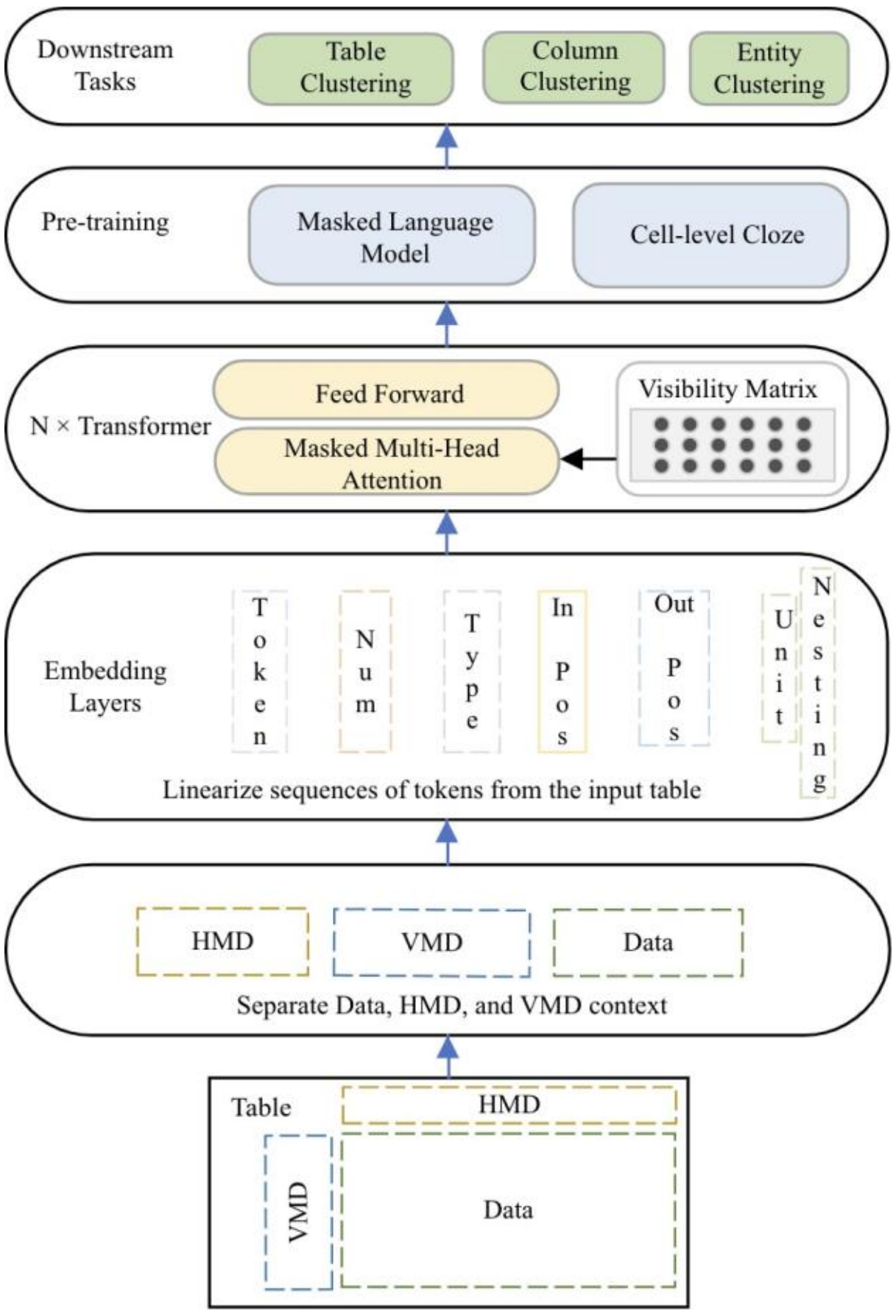

**Figure 2: Transformer-based Deep-learning Architecture with 6 Embeddings Layers for non-1st Normal Form Table with Hierarchical Vertical, Horizontal Metadata, and Nesting.**

The architecture diagram is illustrated in Figure 2[1]. Tabbie [36] uses two separate transformer models [22] to encode a table *row-wise* and *column-wise* separately, then aggregates the representations obtained from both encoders. We segment a table into *three* distinct parts: data, HMD, and VMD. We then concatenate the embeddings from each segment into a composite embedding vector to capture a comprehensive representation of the entire table. This segmentation ensures that the *context* from these semantically different table segments is treated separately by the model and that unique structural and semantic characteristics of each segment are preserved. E.g., *hierarchical* metadata is expected to have hierarchical relationships between the neighbors, encoded/learned in their contextual pattern; neighbors in the data segment might systematically belong to the same domain [18] or represent different properties of the same object [18]. These distinct patterns should be learned independently to maximize accuracy and minimize the training set sizes [13, 69]. We and other researchers studied separating metadata and data in context of different relational data classification tasks and observed that it generally improves performance [7, 13, 41, 57, 65, 69]. All the above-mentioned rationale served as a basis for table segmentation and separate training.

[1]Figure 2 should be read bottom to top.

## 3.1 Embedding Layer

We introduced 6 *new embeddings* into the embedding layer of our new transformer-based architecture explicitly encoding the bi-dimensional coordinates, semantic information about the entity type (for strings), units (for numerical data), and nested tables. We partition the tables into three segments — data, HMD, and VMD and process them separately to separate contexts for each of these types of data that carry different semantics. We iterate over the table cells row by row to train our data row model. We iterate over the table cells column by column to train our data column model. We tokenize cells using [22], embed tokens jointly and create 6 new embeddings corresponding to: token semantics $E_{tok}$, numerical properties $E_{num}$, in-cell position $E_{c_{pos}}$, in-table position $E_{t_{pos}}$, cell features $E_{fmt}$, and inferred type $E_{type}$.

**Token.** To learn token semantics, we use the vocabulary V defined in [45]. The numbers are tokenized using the special token [VAL] (as indicated in "Token" column of Figure 3). The trainable embedding weight for each token is defined by $W_{tok} \in \mathbb{R}^{H \times V}$. The trainable embedding for a token is defined as:

$$E_{tok} = W_{tok} \cdot x_{tok} \tag{2}$$

where $x_{tok}$ is the index of the token in $V$.

**Number.** The numbers are encoded in our embedding vectors using four discrete features, magnitude $x_{mag} \in [0, M]$, precision $x_{pre} \in [0, P]$, the first digit $x_{fst} \in [0, F]$, and the last digit $x_{lst} \in [0, L]$ as in [80]. These features are then one-hot encoded. For example, number 20.3 in Figure 3 is encoded as $(x_{mag}, x_{pre}, x_{fst}, x_{lst}) \rightarrow (2, 2, 2, 3)$. The weights $W_{mag}, W_{pre}, W_{fst}$, and $W_{lst} \in \mathbb{R}^{M/P/F/L \times \frac{H}{4}}$ are concatenated. $M, P, F, L = 10$. The final trainable embedding for the numerical properties is:

$$E_{num} = E_{num_{mag}} \oplus E_{num_{pre}} \oplus E_{num_{fst}} \oplus E_{num_{lst}} \tag{3}$$

where $E_{num_{mag}} = W_{mag} \cdot x_{mag}$, $E_{num_{pre}} = W_{pre} \cdot x_{pre}$, $E_{num_{fst}} = W_{fst} \cdot x_{fst}$ and $E_{num_{lst}} = W_{lst} \cdot x_{lst}$, and $\oplus$ denotes vector concatenation operator.

**In-position.** The in-cell position refers to the index of a token within a cell (Figure 3 "In Pos" column). To represent each position, we introduce a trainable embedding $E_{c_{pos}}$ [80] denoted as:

$$E_{c_{pos}} = W_{c_{pos}} \cdot x_{c_{pos}} \tag{4}$$

where $W_{c_{pos}} \in \mathbb{R}^{H \times I}$ represents the learnable weight, $x_{c_{pos}}$ is the the one-hot encoded position, $I = 64$ is the pre-defined maximum allowable number of tokens within a cell. We trim tokens in each cell where the length exceeds this limit.

**Out-position.** "Out pos" column in Figure 3 is comprised of two components. The first one corresponds to the Bi-dimensional coordinates of the cell, and the second one corresponds to the cell coordinate in the nested table. The nested position embedding incorporates the new spatial coordinate (x, y) for tokens in the nested cell starting with index 1. In the context of a relational table without nesting, our bi-dimensional coordinates reduce to the standard Cartesian coordinates. For cells without nesting the default coordinate (0,0) is used. We randomly initialize the weights for these positional embedding and train them jointly with the attention layers as in [22, 80]. Finally, we concatenate the Bi-dimensional coordinate embedding and nesting coordinate embedding to get the final composite positional embedding.

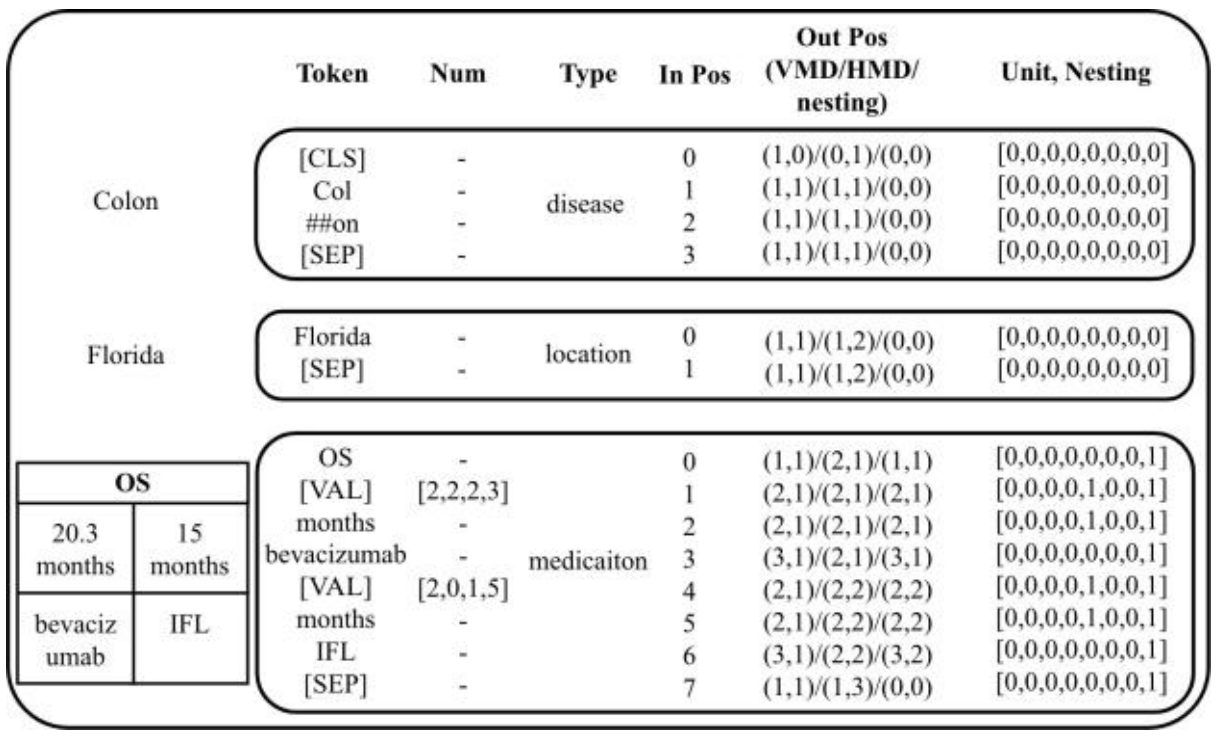

**Figure 3: The Encoded Representation of Table 1 in the Embedding Layer.**

$$E_{t_{pos}} = E_{t_{vpos}} \oplus E_{t_{hpos}} \oplus E_{t_{npos}} \tag{5}$$

where $E_{t_{vpos}} = W_{v_r} \cdot x_{v_r} \oplus W_{v_c} \cdot x_{v_c}$ is the composite embedding for the vertical metadata coordinate position, $E_{t_{hpos}} = W_{h_r} \cdot x_{h_r} \oplus W_{h_c} \cdot x_{h_c}$ is the composite embedding for the horizontal metadata coordinate position, $E_{t_{npos}} = W_{n_r} \cdot x_{n_r} \oplus W_{n_c} \cdot x_{n_c}$ is the composite embedding for the nested coordinate position, $x_{v_r}$ , $x_{v_c}$ , $x_{h_r}$, $x_{h_c}$, $x_{n_r}$, $x_{n_c}$ are the one-hot encoded positions indicating the row and column indexes for each vertical, horizontal, and nested coordinate, $W_{v_r}, W_{v_c}, W_{h_r}, W_{h_c}, W_{n_r}, W_{n_c} \in \mathrm{R}^{G \times \frac{H}{6}}$ are the embedding weights for the vertical metadata row, vertical metadata column, horizontal metadata row, horizontal metadata column, nested row and column positions. $G$ is the maximum number of tuples in a table. We have found $G = 256$ to be sufficient for our datasets.

**Units and Nesting.** To account for the presence of *units* together with numbers and nesting cells we encode them in our last ($6^{th}$ embedding vector) in Figure 3 as one-hot 8-dimensional encoded binary feature vector ("Unit, Nesting" column). The order of one-hot encoding for *units* and *nesting* is [stats, length, weight, capacity, time, temperature, pressure, *nested*], 'stats' indicates statistical measure such as percentage, mean, gaussian etc. The first seven bits in the vector represent the unit. We populate them only for numerical values. The last bit indicates the presence of a nested table in the cell. The embedding for the nested cells coordinate is incorporated in the "Out Pos" component of the embedding layer discussed above. We get the cell features embedding, representing units and nesting, by transforming the feature vector $x \in \mathrm{B}^F$ into the vector space of dimensionality $H$ with weight $W_{fmt} \in \mathrm{R}^{F \times H}$ and bias $b \in \mathrm{R}^H$. $W_{fmt}$ and $b$ are learned during the pre-training phase.

$$E_{fmt} = W_{fmt} \cdot x + b \tag{6}$$

In our case F = 8 is the number of our cell features.

**Type Inference.** We use [60] for type inference and tagging chemicals, diseases, medication types, drugs, etc. On top of this we also defined a custom list of named-entities, types, and noun-phrases for our datasets, such as *vaccines*, *treatments*, *therapies*, *prescriptions* that are beyond capabilities (too domain-specific) of the SOTA NLP packages, when applied to CovidKG and CancerKG datasets. For generic entities such as *name*, *places*, *measurement* we used the *en_core_web_sm* pipeline package for English [26]. In addition, we tag numeric, range, and text types using standard *regex* in Python. The embedding for type inference is of size (14, 768). 768 is the dimensionality of the hidden layer of our model and 14 is the number of different supported types in our experiment. All tokens in a cell get the same type. For example, in Figure 3, tokens corresponding to the cell "colon" are typed as disease. The type inference mapping has a finite set of size $T = 14$. Each token is assigned with a trainable embedding in $W_{type} \in \mathrm{R}^{H \times T}$.

$$E_{type} = W_{type} \cdot x_{type} \tag{7}$$

The final embedding vector of a token is the summation of all the components:

$$E = E_{tok} + E_{num} + E_{c_{pos}} + E_{t_{pos}} + E_{type} + E_{fmt} \tag{8}$$

**Table 2: A sample Relational Table.**

| *Name* | *Age* | *Job* |
|---|---|---|
| Sam | 24 | Engineer |
| John | 25 | Scientist |
| Nick | 23 | Lawyer |

## 3.2 Visibility Matrix

We introduce a custom visibility matrix to make the attention mechanism attend only to the neighboring structural context of the same kind (i.e., carrying the same semantics), thus avoiding redundant information. The standard self-attention mechanism allows every token in a table to attend to every other token, regardless of where the tokens are – in the cell, in the same *tuple*, *column* or one in the *data* cell another in the *metadata*. Spatial information is valuable as it is representative of separate segments of a table carrying different semantics (HMD, VMD, D). Hence it is important to precisely capture and encode it, which we accomplish through our visibility matrix. The standard transformer attention mechanism is also capable of capturing it, but our visibility matrix makes it more explicit [14, 21, 36, 80]. Consider an example Table 2: 'Sam' and 'Engineer' are related because they are in the same row, whereas 'Sam' should not be related to 'Lawyer'. Similarly, 'Scientist' is related to '*Job*', but should not be related to the attribute '*Age*'. To accurately model this important structural information in tables, we must have a mechanism to explicitly inform the model about which token/cells are *structurally* related. This is achieved by introducing an attention mask or as we call it - *visibility matrix*. An experiment in our ablation study in Section 4.6, where we remove the *visibility matrix* (thus resort to the standard attention mechanism), demonstrates that it results in a substantial loss in accuracy, hence justifies its value.

Our visibility matrix is a binary matrix used as an attention mask in the transformer layer during calculation of a multi-head self-attention. Table cells in the same row or column are visible to each other, i.e., if element $i$ is a token in a table and if element $j$ is a token in same row or column, $M_{ij} = 1$. $M_{ij} = 1$ if and only if element $i$ is visible to element $j$, otherwise $M_{ij} = 0$. An element here can be a token in the header or data cell. We apply the same visibility matrix separately to data, vertical, and horizontal metadata, hence treating these semantically different context types separately, unlike other SOTA solutions [4, 12, 25, 33, 36, 80, 87].

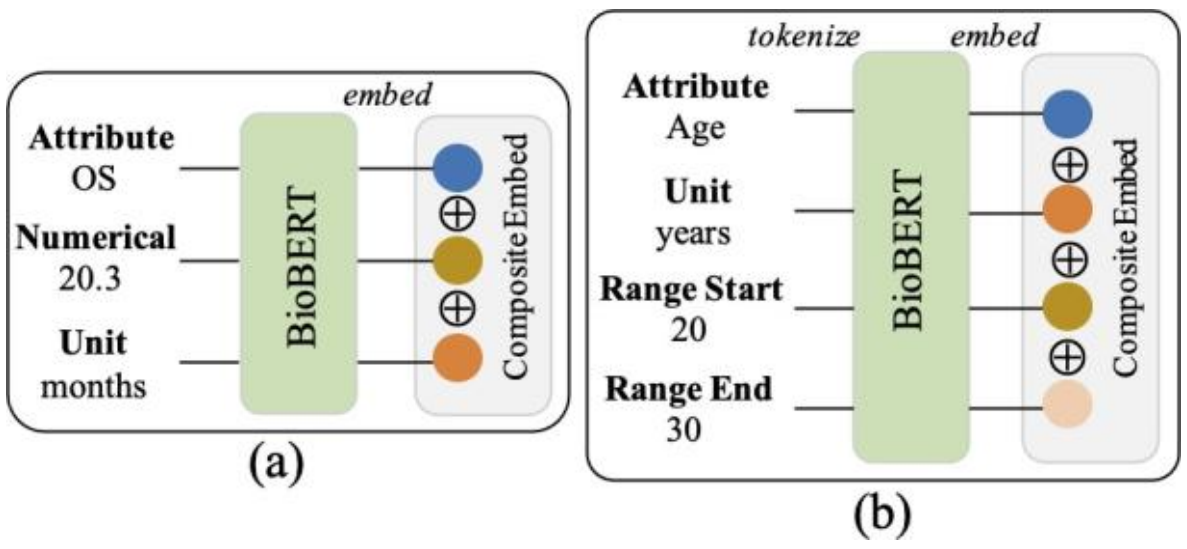

**Figure 4: Composite Embedding (CE) Structure for (a) Numerical Attributes and (b) Ranges.**

## 3.3 Pre-training Methodology

We took the vocabulary and pre-trained token embeddings and encoder weights from BioBERT [45] to initialize TabBiN for pre-training on our 5 datasets. We trained each version of our model for 50,000 steps, batch size 12, learning rate 2e-5. We trained 4 models – 2 for data – tuples, columns; 2 for metadata – horizontal, vertical metadata. While reading a row or a column and generating the training sets, we are keeping track of the respective Bi-dimensional coordinates for each cell so that we can include the positional information in our embeddings (see Figure 2). We add [*CLS*] at the start of each row/column and [*SEP*] between the cells. We use table sequences with no more than 256 tokens that we found to be sufficient for our datasets (i.e., increases beyond 256 prolong the fine-tuning process, without increasing accuracy on our downstream tasks). We use the *Masked Language modeling* and *Cell-level cloze* as our training objectives [22, 45, 80]. We separate the model pre-training for data and metadata, so their context is treated separately. For example, in TabBiN data column model we pre-trained the model to learn the columnar data context, excluding metadata. We used AWS p3.2xlarge instances. Pre-training of each model took approximately five hours.

## 3.4 Composite Embeddings (CE)

For using BioBERT embeddings for numerical values we came up with the idea to have composite structure concatenating (⊕) embeddings for the attribute, its value and the unit. Figure 4(a) illustrates this process for a column "OS" (i.e., Overall Survival) from a nested table in Table 1, attribute "OS" has numerical value "20.3 *months*". This structure preserves the actual meaning of the numerical value together with the unit. The composite embedding for *Range* values has similar structure, where we concatenate the embeddings for the attribute, unit, range start, range end. In Figure 4(b) we show this structure with an example attribute *Age* having the numerical range "20-30" and the unit "*year*".

# 4 EXPERIMENTAL EVALUATION

We evaluated our TabBiN embeddings on 3 popular downstream tasks – Column Clustering (CC), Table Clustering (TC) by topic, Entity Clustering (EC). We performed our evaluation on 5 large-scale datasets described above in section 2.2 – Webtables [46], CovidKG [56], CancerKG, CIUS [30, 80] and SAUS [30, 80]. To compare against the SOTA transformer-based model supporting structured data we fine-tuned TUTA [80]. We also fine-tuned one of the top transformer-based models for biomedical data – BioBERT [45], classic Word2Vec [58] embeddings model, and DITTO [49] entity matching model on our data sets.

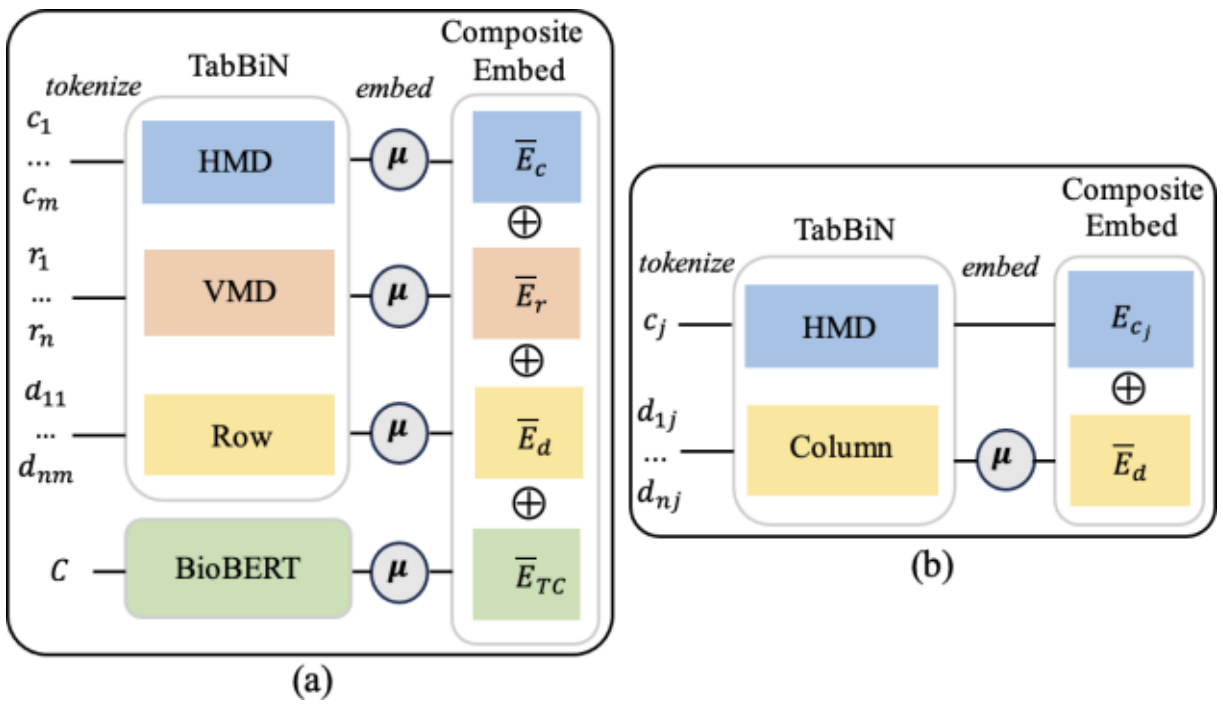

**Figure 5: Composite Embedding (CE) for (a) Table Clustering (TC) and (b) Column Clustering (CC).**

**Table 3: The Average Training time vs. MAP/MRR for CC and TC tasks on CancerKG (tables with string data) for different dimensionality of Word2vec embeddings.**

| *Dimensionality* | *Training Time(hours)* | *CC* | *TC* |
|---|---|---|---|
| 100 | 2.5 | 0.50/0.60 | 0.55/0.45 |
| 200 | 4 | 0.45/0.60 | 0.65/0.46 |
| 300 | 6 | 0.60/0.65 | 0.70/0.50 |
| 400 | 7 | 0.59/0.65 | 0.70/0.50 |
| 1024 | 7.5 | 0.60/0.65 | 0.70/0.50 |

**TUTA.** We download the pre-trained TUTA explicit model and fine-tune it on our datasets using identical hyper-parameters to those of TabBiN. We tokenize, embed, and encode each table as described in [80]. Training took ≈ 4.5 hours on AWS p3.2xlarge instance.

**BioBERT.** We fine-tune the original BioBERT for 50K steps, batch size 12, learning rate 2e-5, on a Linux server with 80 Intel Xeon cores, 256 GB RAM for ≈ 41 days. The training set is comprised of table tuples. We also fine-tuned a second BioBERT model including table captions as the embedding vector component (see Figure 5(a), Table 11).

**Word2vec.** We train Word2vec model with embedding dimensionality 300, the context window of size 3 before and after the target word, minimum count of 1 for word inclusion. We did experiments with several embedding dimensions as shown in Table 3 and found no notable performance difference when using the embeddings trained with the dimension more than 300. However, the slowdown in training time was significant so we chose 300 as optimal dimensionality. We trained Word2Vec on table tuples on AWS p3.2xlarge instance.

**DITTO.** The downstream task that Ditto is built for is entity classification, where *entity is a tuple*. Ditto performs *binary classification* to decide on a match or mismatch, whereas we compute cosine similarity between each entity and sort in descending order to get a cluster of matched entities. In order to compare to Ditto, we added a *linear layer* followed by *softmax layer* on top of our TabBiN transformer layers, and an ensemble, so TabBiN can also perform binary classification. We have included the additional experiments comparing to Ditto both on ours and Ditto's datasets in Table 9. We use AP@20 to evaluate the quality of our formed entity clusters. We train DITTO using RoBERTa [55] pre-trained model and default hyperparameters mentioned in

**Table 4: MAP/MRR for CC – Textual and Numerical.**

| | *Datasets* | *TabBiN* | *TUTA* | *BioBERT* | *Word2vec* |
|---|---|---|---|---|---|
| *All tables* | CancerKG | **0.90 / 1.00** | 0.70 /0.95 | 0.80 / 0.92 | 0.60 / 0.65 |
| | CovidKG | **0.90 / 1.00** | 0.80 /**1.00** | 0.80 / 0.95 | 0.60 / 0.70 |
| | Webtables | **0.95 / 0.90** | 0.90 /0.80 | 0.85 / 0.86 | 0.40 / 0.60 |
| | CIUS | **0.90 / 0.95** | **0.90** /0.92 | 0.80 / 0.90 | 0.65 / 0.50 |
| | SAUS | **0.80 / 0.95** | 0.65 /**0.95** | 0.70 / 0.90 | 0.42 / 0.50 |
| *Small tables* | CancerKG | **0.95 / 0.95** | 0.77 /**0.95** | 0.81 / **0.95** | 0.54 / 0.70 |
| | CovidKG | **0.80 / 0.98** | 0.66 /0.95 | 0.78 / 0.88 | 0.56 / 0.65 |
| | Webtables | **0.60 / 1.00** | 0.50 /**1.00** | **0.60 / 1.00** | 0.40 / 0.80 |
| | CIUS | **0.98 / 0.98** | 0.90 /**0.98** | 0.81 / 0.90 | 0.60 / 0.52 |
| | SAUS | **0.90 / 0.98** | 0.85 /0.95 | 0.81 / 0.92 | 0.67 / 0.50 |
| *Large tables* | CancerKG | **0.90 / 1.00** | 0.72 /0.98 | 0.85 / 0.90 | 0.60 / 0.65 |
| | CovidKG | **0.85 / 1.00** | 0.64 /**1.00** | 0.68 / 0.90 | 0.53 / 0.65 |
| | Webtables | **0.98 / 0.90** | 0.96 /**0.90** | 0.85 / **0.90** | 0.53 / 0.70 |
| | CIUS | **0.96 / 1.00** | 0.90 /0.95 | 0.77 / 0.90 | 0.67 / 0.50 |
| | SAUS | **0.81 / 0.98** | 0.78 /0.95 | 0.78 / 0.92 | 0.40 / 0.50 |
| *Numerical* | CancerKG | **0.80 / 0.95** | 0.60 /0.92 | 0.72 / 0.92 | 0.25 / 0.50 |
| | CovidKG | **0.60 / 0.90** | 0.50 /0.85 | **0.60 / 0.90** | 0.20 / 0.40 |
| | Webtables | **0.78 / 1.00** | 0.50 /0.90 | 0.50 / 0.95 | 0.20 / 0.45 |
| | CIUS | **0.80 / 0.90** | 0.74 /**0.90** | **0.80** / 0.85 | 0.40 / 0.40 |
| | SAUS | **0.78 / 0.90** | 0.72 /**0.90** | 0.70 / 0.85 | 0.15 / 0.48 |
| *Ranges* | CancerKG | **0.97 / 1.00** | 0.94 /**1.00** | 0.80 / **1.00** | 0.70 / 0.80 |
| | SAUS | **0.98 / 1.00** | **0.98 /1.00** | 0.77 / **1.00** | 0.64 / 0.80 |

[49]. We created five different labeled training datasets consisting of positive and negative pairs of matching and non-matching entities from entity types corresponding to each dataset defined in Table 7. For CancerKG and CovidKG we have 5*k* positive and 5*k* negatively labeled pairs. For Webtables we have 1.5*k* positive and 1.5*k* negatively labeled pairs and for each CIUS and SAUS we have 400 positive and 400 negatively labeled pairs. The average training time for DITTO is ≈ 3.2 hours.

As our evaluation measures, we use Mean Average Precision [52] (MAP@20) and Mean Reciprocal Rank [20] (MRR@20) calculated on the sorted list of clustered columns, tables, or entities (by *cosine* similarity in the descending order). We compute AP@20 and average it over a sample of different columns, tables and entities from each dataset and report it in Tables 3-8 with the best results indicated in boldface. For comparison against DITTO entity matching, results are measured using F1 score.

## 4.1 Column Clustering (CC)

For CC we create a composite embedding by concatenating the embedding $E_{c_j}$ for an attribute $c_j$ in HMD from our TabBiN-HMD model (i.e., trained only on HMD) and the average ($\mu$) embedding $\bar{E}_d$ over data cell tokens for corresponding $c_j$ column from our TabBiN-column model (i.e., trained only on columns) as shown in Figure 5(b). We match two columns by calculating the *cosine* similarity between their TabBiN embedding vectors. We use LSH-based blocking [28] to avoid quadratic complexity for the entire dataset. To cluster columns, for each column, we create a list of similar columns, sorted by the cosine similarity in descending order, the top 20 entries form a cluster. We separate the columns that we have (i.e., 227,279 in CancerKG) into columns having strictly numerical or string values. Table 4 illustrates the experimental results comparing TabBiN to the SOTA models.

**Table 5: MAP/MRR for TC – Tables with HMD versus HMD/VMD, mostly Numerical Content, with Nesting.**

| | *Datasets* | *TabBiN* | *TUTA* | *BioBERT* | *Word2vec* |
|---|---|---|---|---|---|
| *HMD* | CancerKG | **0.87 / 1.00** | 0.78 / **1.00** | 0.67 / **1.00** | 0.53 / **0.90** |
| | CovidKG | **0.78 / 0.95** | 0.64 / 0.90 | 0.60 / 0.90 | 0.40 / 0.85 |
| | Webtables | **0.87 / 1.00** | 0.81 / 0.98 | 0.80 / 0.95 | 0.40 / 0.88 |
| | CIUS | **0.50 / 0.90** | 0.40 / **0.90** | 0.40 / **0.90** | 0.10 / 0.40 |
| | SAUS | **0.50 / 0.90** | 0.40 / **0.90** | 0.40 / **0.90** | 0.10 / 0.40 |
| *HMD+VMD* | CancerKG | **0.80 / 0.92** | 0.70 / 0.85 | 0.68 / 0.80 | 0.10 / 0.40 |
| | CovidKG | **0.85 / 0.95** | 0.80 / **0.95** | 0.70 / 0.80 | 0.15 / 0.45 |
| | Webtables | **0.90 / 1.00** | 0.84 / 0.98 | 0.80 / 0.85 | 0.20 / 0.45 |
| | CIUS | **0.54 / 0.95** | 0.53 / 0.90 | 0.40 / 0.75 | 0.10 / 0.35 |
| | SAUS | **0.54 / 0.95** | **0.54** / 0.85 | 0.40 / 0.78 | 0.10 / 0.35 |
| *> 80% Num* | CancerKG | **0.81 / 0.90** | 0.70 / 0.85 | 0.60 / 0.80 | 0.10 / 0.38 |
| | CovidKG | **0.53 / 0.90** | 0.30 / 0.80 | 0.50 / 0.82 | 0.18 / 0.35 |
| | Webtables | **0.67 / 0.95** | 0.58 / 0.85 | 0.58 / 0.80 | 0.10 / 0.30 |
| | CIUS | **0.40 / 0.90** | 0.30 / 0.82 | 0.30 / 0.80 | 0.10 / 0.36 |
| | SAUS | **0.41 / 0.90** | 0.30 / 0.82 | 0.32 / 0.80 | 0.10 / 0.36 |
| *Nesting* | CancerKG | **0.85 / 1.00** | 0.68 / 0.80 | 0.60 / 0.75 | 0.20 / 0.42 |
| | CovidKG | **0.70 / 0.95** | 0.60 / 0.80 | 0.54 / 0.70 | 0.18 / 0.38 |

## 4.2 Table Clustering (TC)

Similarly for TC we create the composite embedding by concatenating the average embedding $\bar{E}_d$ for data cells from the TabBiN-row model, the average embedding $\bar{E}_c$ for HMD from our TabBiN-HMD model, the average embedding $\bar{E}_r$ for VMD from our TabBiN-VMD model, and the average embedding $\bar{E}_{TC}$ for the table caption taken from the BioBERT model fine-tuned on our datasets as illustrated in Figure 5(a). We use *cosine* similarity as a distance measure between our TabBiN embedding vectors corresponding to the tables to form cohesive clusters. To form clusters, we first calculate a centroid embedding vector for a given topic table. Then, we compute distance from other tables to this centroid vector, sorted in descending order to form the cluster with top 20 entries. We did it for centroids corresponding to different topics and report the MAP/MRR@20.

## 4.3 Entity Clustering (EC)

We took sets of columns with labels specific to our datasets (i.e., *drugs*, *vaccines*, *symptoms*, *diseases*, *crime*, *states*, *cities*, etc.) and extracted their corresponding data values. This approach resulted in very large and high-quality catalogs of entities, both domain-specific (i.e., CancerKG, CovidKG) as well as more generic (i.e., Webtables). Evaluation of these catalogs is reported in Table 7. For each dataset the average precision (AP) was calculated by taking a sample of size 40 and having two annotators label them. Next, we selected entities of each of 18 entity types that we work with in each dataset (e.g., drugs) and calculated the cosine similarity between each entity and the remaining entities in the dataset, sorted in descending order, calculated AP@20 for each cluster (formed by taking top 20 entities) corresponding to an entity type and averaged it. We used TabBiN-column model for this EC task. The average F1 measure of 5 runs is reported in Table 9.

**Table 6: MAP/MRR for TC – Tables with Relational versus Non-relational. Heterogeneous Data Types.**

| | *Datasets* | *TabBiN* | *TUTA* | *BioBERT* | *Word2vec* |
|---|---|---|---|---|---|
| *Relational* | CancerKG | 0.92 / **1.00** | **0.94 / 1.00** | 0.80 / 0.80 | 0.70 / 0.65 |
| | CovidKG | **0.80 / 0.90** | 0.72 / 0.85 | 0.75 / 0.65 | 0.40 / 0.60 |
| | Webtables | **0.84 / 1.00** | 0.77 / **1.00** | 0.70 / 0.80 | 0.20 / 0.50 |
| | CIUS | **0.42 / 0.90** | 0.40 / **0.90** | 0.35 / **0.90** | 0.15 / 0.80 |
| | SAUS | **0.50 / 0.92** | 0.45 / 0.90 | 0.40 / 0.80 | 0.10 / 0.65 |
| *Non-Relational* | CancerKG | **0.77 / 0.88** | 0.71 / 0.80 | 0.40 / 0.70 | 0.10 / 0.30 |
| | CovidKG | **0.74 / 0.90** | 0.70 / **0.90** | 0.40 / 0.70 | 0.10 / 0.30 |
| | Webtables | **0.90 / 0.90** | 0.85 / 0.85 | 0.70 / 0.85 | 0.10 / 0.35 |
| | CIUS | **0.40 / 0.90** | **0.40 / 0.90** | 0.32 / 0.80 | 0.10 / 0.60 |
| | SAUS | **0.46 / 0.90** | 0.40 / **0.90** | 0.30 / 0.80 | 0.10 / 0.60 |
| *String* | CancerKG | **0.92** / 0.98 | **0.92 / 1.00** | 0.80 / 0.85 | 0.70 / 0.50 |
| | CovidKG | **0.90 / 1.00** | 0.84 / **1.00** | 0.79 / 0.80 | 0.50 / 0.50 |
| | Webtables | 0.84 / **0.95** | 0.70 / 0.90 | 0.68 / 0.70 | 0.40 / 0.48 |
| *Text/Num (50%)* | CancerKG | **0.86 / 1.00** | 0.81 / 0.95 | 0.64 / 0.90 | 0.46 / 0.40 |
| | CovidKG | **0.85 / 0.90** | 0.80 / **0.90** | 0.70 / 0.72 | 0.15 / 0.30 |
| | Webtables | **0.95 / 1.00** | 0.92 / **1.00** | 0.90 / **1.00** | 0.20 / 0.40 |

**Table 7: Entity Catalogs.**

| *Datasets* | *Entity Types* | *Count* | *AP* |
|---|---|---|---|
| CancerKG | drug, therapy. segment, tumor, reagent | 12,553 | 0.72 |
| CovidKG | characteristics, vaccines, symptoms, diseases, infections | 12,573 | 0.85 |
| Webtables | country, company, genre, title, size | 3,316 | 0.796 |
| CIUS | crime, city | 474 | 0.9 |
| SAUS | city, industrial | 507 | 0.9 |

**Table 8: MAP/MRR for EC.**

| *Datasets* | *TabBiN* | *TUTA* | *BioBERT* | *Word2vec* |
|---|---|---|---|---|
| CancerKG | **0.96 / 1.00** | 0.90 / **1.00** | 0.90 / 0.90 | 0.80 / 0.60 |
| CovidKG | **0.94 / 1.00** | 0.90 / **1.00** | 0.88 / 0.90 | 0.72 / 0.50 |
| Webtables | **0.80 / 0.98** | 0.79 / **0.98** | 0.73 / 0.85 | 0.65 / 0.56 |
| CIUS | **0.96 / 1.00** | **0.96 / 1.00** | 0.90 / 0.95 | 0.70 / 0.55 |
| SAUS | **0.96 / 1.00** | 0.90 / **1.00** | 0.88 / 0.90 | 0.70 / 0.60 |

## 4.4 TabBiN Performance Highlights

Column Clustering (CC, Table 4): TabBiN outperforms both TUTA and BioBERT SOTA models on numerical CC task on Webtables with a significant MAP delta of 0.28. Also, TabBiN outperforms BioBERT on large tables by a significant MAP delta of 0.17 on CovidKG. For small tables TabBiN again outperforms BioBERT with a large MAP delta 0.14 on CancerKG. The highest CC MAP of TabBiN is 0.98 and it is achieved on large tables from Webtables, small tables from CIUS, and ranges from SAUS.
Table Clustering (TC, Table 5, Table 6): TabBiN outperforms TUTA on nested table clustering with a significant MAP delta of 0.17 on CancerKG. On tables with HMD from CovidKG TabBiN outperforms TUTA with a large MAP margin of 0.14. TabBiN outperforms TUTA by a large MAP delta of 0.14 on Webtables with string data. TabBiN achieves the highest TC MAP of 0.95 on Webtables with mixed data. On relational tables from CancerKG, TUTA outperforms us in-significantly, with MAP delta of 0.2.
Entity Clustering (EC, Table 8): In Table 8, we can see that TabBiN attains the highest MAP across all datasets for EC. TabBiN outperforms TUTA by a small MAP margin of 0.06 for both CancerKG and SAUS respectively. On entity matching (to compare to DITTO, Table 9), TabBiN outperforms Ditto with a small F1 score margin of 1.92%. on structured Amazon-Google dataset. Ditto outperforms TabBiN on Abt-Buy dataset by a small margin of 1.21%. Similarly, on our datasets Ditto insignificantly outperforms TabBiN by 1.24% and 0.37% deltas in F1 measure.

## 4.5 Composite Embeddings Analysis

We employed separate composite embedding vectors for CC and TC tasks, as illustrated in Figure 5 earlier. We use the following abbreviations for composite embeddings in Tables 10 and 11: *TabBiN-colcomp* for composite embeddings formed by concatenating the embeddings from TabBiN-column model and TabBiN-HMD model; *TabBiN-tblcomp1* for composite embeddings formed by concatenating the embeddings from TabBiN-row model, TabBiN-HMD model and TabBiN-VMD model; *TabBiN-tblcomp2* for composite embeddings formed by concatenating the embeddings from TabBiN-row model, TabBiN-HMD model, TabBiN-VMD model and fine-tuned BioBERT on table captions.
**Column Clustering.** From Table 10, we can conclude that on both numerical and textual tabular data TabBiN composite embeddings perform the best. This is observed on all evaluation datasets. Specifically, for numeric ranges, composite embeddings demonstrate superior performance, as evident on CancerKG. Moreover, on large tables with textual data, composite embeddings excel in performance as observed on two large-scale datasets.
**Table Clustering.** From Table 11 we can conclude that on tables with nesting, tables only with HMD, tables with both HMD and VMD (non-relational tables) and relational tables our composite embeddings perform the best. This is observed on two large-scale datasets and three different types of composite embeddings.

## 4.6 Ablation Studies

We conduct four ablation studies ($TabBiN_{1-4}$ below) to demonstrate the efficiency of our visibility matrix, type inference, units and nesting, and bi-dimensional coordinates. For each ablation study, we train the models removing the corresponding target embedding component and then perform TC and CC evaluation tasks on our datasets. Table 12, 13 illustrate the results.
$TabBiN_1$. Removing our visibility matrix makes TabBiN resort to the standard transformer attention mechanism. We observe that this leads to a *substantial* MAP/MRR drop on all datasets. We observe a drop in MAP for 0.34 on TC on Webtables with string data; for 0.30 on relational Webtables. The drop is more than 0.2 on most of the remaining datasets. For CC the MAP drop is by 0.25 for columns with string data (CancerKG, Webtables) and for 0.23 for numerical columns (CancerKG).
$TabBiN_2$. Without *type inference* CC MAP on columns with string data in CancerKG, Webtables, and SAUS drops by 0.1. For TC in relational (Webtables), non-relational (CancerKG), and Webtables with string data MAP drops by 0.15.
$TabBiN_3$. Removing *Units* and *Nesting* embedding components

Table 9: F1 scores (%) for Entity Classification on ER-Magellan EM datasets [43] and our datasets.

| *Methods* | *Structured Amazon-Google* | *Textual Abt-Buy* | *Dirty Walmart-Amazon* | *CancerKG* | *CovidKG* | *Webtables* | *CIUS* | *SAUS* |
|---|---|---|---|---|---|---|---|---|
| *TabBiN* | **77.50** | 88.12 | **86.06** | **90.7** | **90.46** | 83.50 | **90.48** | 88.84 |
| *DITTO* | 75.58 | **89.33** | 85.69 | 90.2 | 89.29 | **84.74** | 88.78 | **89.21** |

Table 10: MAP/MRR for CC Performance by TabBiN without and with Composite Embeddings.

| | *String Values (any #tuples)* | | *Numeric Values* | | *String (#tuples < 10)* | | *String (#tuples > 10)* | | *Ranges* |
|---|---|---|---|---|---|---|---|---|---|
| | *CancerKG* | *CovidKG* | *CancerKG* | *CovidKG* | *CancerKG* | *CovidKG* | *CancerKG* | *CovidKG* | *CancerKG* |
| *TabBiN-column* | 0.88 / 0.98 | 0.88 / 0.98 | 0.60 / 0.90 | 0.42 / **0.90** | 0.90 / **0.95** | 0.74 / 0.95 | **0.90 / 1.00** | 0.77 / 0.90 | 0.90 / **1.00** |
| *TabBiN-colcomp* | **0.90 / 1.00** | **0.90 / 1.00** | **0.80 / 0.95** | **0.60 / 0.90** | **0.95 / 0.95** | **0.80 / 0.98** | **0.90 / 1.00** | **0.85 / 1.00** | **0.97 / 1.00** |

Table 11: MAP/MRR for TC Performance by TabBiN without and with Composite Embeddings – Tables with Heterogeneous Data, Nesting, HMD versus HMD and VMD, Relational.

| | *String Values* | | *Text/Num(50%)* | | *> 80% Num* | | *Nesting* | | *HMD* | | *HMD+VMD* | | *Relational* | |
|---|---|---|---|---|---|---|---|---|---|---|---|---|---|---|
| | *CancerKG* | *CovidKG* | *CancerKG* | *CovidKG* | *CancerKG* | *CovidKG* | *CancerKG* | *CovidKG* | *CancerKG* | *CovidKG* | *CancerKG* | *CovidKG* | *CancerKG* | *CovidKG* |
| *TabBiN-row* | 0.82/ 0.90 | 0.80/ 0.90 | 0.80/ 0.90 | 0.77/ 0.85 | 0.70/ 0.80 | 0.40/ 0.90 | 0.72/ 0.95 | 0.65/ 0.90 | 0.84/ 0.95 | 0.7/ 0.90 | 0.77/ 0.90 | 0.72/ 0.90 | 0.90/ **1.00** | 0.72/ **0.90** |
| *TabBiN-tblcomp1* | 0.88/ 0.90 | 0.85/ 0.92 | 0.80/ 0.95 | 0.80/ 0.90 | 0.74/ 0.80 | 0.40/ 0.90 | 0.8/ **1.00** | 0.68/ 0.90 | 0.84/ 0.95 | 0.72/ **0.95** | 0.77/ 0.90 | 0.76/ 0.90 | 0.90/ **1.00** | 0.74/ **0.90** |
| *TabBiN-tblcomp2* | **0.92/ 0.98** | **0.90/ 1.00** | **0.86/ 1.00** | **0.85/ 0.90** | **0.81/ 0.90** | **0.53/ 0.90** | **0.85/ 1.00** | **0.70/ 0.95** | **0.87/ 1.00** | **0.78/ 0.95** | **0.80/ 0.92** | **0.85/ 0.95** | **0.92/ 1.00** | **0.80/ 0.90** |

decreases MAP on nested tables (CancerKG) by 0.25. There is 0.22 decrease in MAP on numerical Webtables. For numerical columns on CC the drop in MAP is 0.21 (CancerKG). We can see a notable decrease in MAP for both CC and TC tasks in other datasets too.

$TabBiN_4$. Removing our bi-dimensional coordinates erases the explicit encoding of the positions of all data and metadata cells in two dimensions in the main table as well as in the nested in-cell tables (e.g., in Figure 1). Our nesting definition includes tables nested inside a cell having their own attributes (e.g., in Figure 1), which is different from the classical notion of nesting/unnesting. The removal leads to a significant drop in MAP on CC for both numerical and string columns in CancerKG by 0.12 and 0.11 respectively. Similarly on TC, MAP for nested tables (CancerKG) drops by 0.15, MAP for numerical tables (>80% Num) in CovidKG drops by 0.13 and MAP for relational tables (CancerKG) drops by 0.12.

We conclude that removing either of the visibility matrix, type inference, units and nesting, or bi-dimensional coordinates significantly hurts TabBiN performance as evidenced by four ablation studies.

Table 12: MAP/MRR for Ablation Study on CC.

| | *Datasets* | *TabBiN* | $TabBiN_1$ | $TabBiN_2$ | $TabBiN_3$ | $TabBiN_4$ |
|---|---|---|---|---|---|---|
| *String* | CancerKG | **0.90/1.00** | 0.65/0.80 | 0.80/**1.00** | 0.70/0.87 | 0.79/0.90 |
| | CovidKG | **0.90/1.00** | 0.74/0.82 | 0.85/**1.00** | 0.74/0.85 | 0.82/0.90 |
| | Webtables | **0.95/0.90** | 0.70/0.85 | 0.85/**0.90** | 0.80/**0.90** | 0.85/0.88 |
| | CIUS | **0.90/0.95** | 0.66/0.80 | 0.86/0.90 | 0.80/0.90 | 0.80/0.85 |
| | SAUS | **0.80/0.95** | 0.58/0.90 | 0.70/0.90 | 0.65/0.80 | 0.70/0.80 |
| *Numerical* | CancerKG | **0.80/0.95** | 0.65/0.80 | 0.84/**0.95** | 0.59/0.78 | 0.68/0.86 |
| | CovidKG | **0.60/0.90** | 0.48/0.80 | **0.60/0.90** | 0.50/0.82 | 0.50/0.80 |
| | Webtables | **0.78/1.00** | 0.60/0.85 | 0.76/**1.00** | 0.65/0.90 | 0.70/0.88 |
| | CIUS | **0.80/0.90** | 0.65/0.78 | **0.80/0.90** | 0.60/0.82 | 0.70/0.85 |
| | SAUS | **0.78/0.90** | 0.65/0.88 | **0.78/0.90** | 0.62/0.80 | 0.67/0.85 |

## 4.7 Large Language Models (LLMs) and Retrieval Augmented Generation (RAG)

Motivated by the ongoing popularity of LLMs, we compared our embeddings on two large-scale datasets (CancerKG and CovidKG) against several major LLMs on two downstream tasks — column and table clustering. We fine-tuned Llama2 [74] and GPT-2 [66] due to their availability in open-source repositories, hence affordability for fine-tuning. We used *llama-2-7b-chat* model, which is a part of a collection of pre-trained and fine-tuned generative text models with 7 billion parameters. We did not fine-tune GPT-3.5 [72] and GPT-4 [54] due to very high cost of doing that at scale of our datasets. For these two models we could only afford to use samples of our datasets for evaluation. For RAG+GPT-3.5 and RAG+GPT-4, however, we first used RAG with an example (i.e., a table or a column) on the entire datasets, so it reduced its size, so it could be ingested into the GPT model via its API for a reasonable cost for further downstream task execution. Finally, we submit prompts to LLMs, requesting to perform our downstream tasks. Following each prompt, we collected and evaluated the models' responses by calculating AP@20 and averaging it. For both tasks we observed lower MAP/MRR for Llama 2 and GPT-2 on both datasets (Table 14). We repeated similar experiments with Retrieval-augmented generation (RAG) to improve the quality of LLMs responses. We have chosen Sycamore [1], a well-known RAG system. We put substantial effort to integrate recent LLMs, such as Llama2, GPT-3.5 and GPT-4 into Sycamore [1] for our experiments. We can see from the results that RAG improves performance. The improvement is significant in case of Llamma2 with RAG (for textual CC on CancerKG MAP increase by 0.30), but falls short of TabBiN. Similarly, we observe increase in MAP values, especially with GPT-4, but again TabBiN outperforms both GPT-3.5 and GPT-4 on our CC and TC downstream task. However, RAG+GPT-4 achieves perfect MRR score (the second metric), outperforming us by a delta of 0.1 (the last two columns in Table 14). This is because MRR only considers a single highest-ranked result [20], and RAG+GPT-4 turns out to be great at providing the first item correctly, while TabBiN sometimes makes mistakes in the first position. TabBiN performs better however when ranking of all relevant items are considered, as captured by the first metric, MAP [52] (Table 14).

Table 13: MAP/MRR for Ablation Study on TC.

| | Datasets | TabBiN | $TabBiN_1$ | $TabBiN_2$ | $TabBiN_3$ | $TabBiN_4$ |
|---|---|---|---|---|---|---|
| Relational | CancerKG | **0.92/1.00** | 0.65/0.90 | 0.85/**1.00** | 0.76/0.95 | 0.80/0.80 |
| | CovidKG | **0.80/0.90** | 0.58/0.80 | 0.70/0.85 | 0.60/0.88 | 0.72/0.82 |
| | Webtables | **0.84/1.00** | 0.54/0.95 | 0.69/0.95 | 0.70/**1.00** | 0.75/**1.00** |
| | CIUS | **0.42/0.90** | 0.35/**0.90** | 0.40/**0.90** | 0.30/0.87 | 0.35/0.88 |
| | SAUS | **0.50/0.92** | 0.35/0.90 | 0.50/0.90 | 0.34/0.90 | 0.42/0.85 |
| Non-Relational | CancerKG | **0.77/0.88** | 0.48/0.80 | 0.62/0.85 | 0.56/0.80 | 0.68/0.80 |
| | CovidKG | **0.74/0.90** | 0.60/0.83 | 0.66/0.88 | 0.60/0.88 | 0.65/0.85 |
| | Webtables | **0.90/0.90** | 0.70/0.86 | 0.80/**0.90** | 0.70/0.85 | 0.80/0.88 |
| | CIUS | **0.40/0.90** | 0.35/0.82 | 0.40/**0.90** | 0.30/0.80 | 0.36/0.83 |
| | SAUS | **0.46/0.90** | 0.40/0.87 | 0.45/**0.90** | 0.30/0.85 | 0.40/0.85 |
| > 80% Num | CancerKG | **0.81/0.90** | 0.60/0.85 | 0.78/**0.90** | 0.63/0.80 | 0.70/0.86 |
| | CovidKG | **0.53/0.90** | 0.32/**0.90** | 0.52/0.86 | 0.35/0.84 | 0.40/0.85 |
| | Webtables | **0.67/0.95** | 0.44/0.90 | 0.65/0.90 | 0.45/0.90 | 0.60/0.92 |
| | CIUS | **0.40/0.90** | 0.35/0.70 | 0.40/**0.90** | 0.35/0.88 | 0.38/**0.90** |
| | SAUS | **0.41/0.90** | 0.35/0.75 | 0.40/**0.90** | 0.31/**0.90** | 0.38/0.88 |
| String | CancerKG | **0.92/0.98** | 0.66/0.90 | 0.88/0.80 | 0.82/0.92 | 0.86/0.90 |
| | CovidKG | **0.90/1.00** | 0.70/0.90 | 0.80/0.85 | 0.90/0.95 | 0.88/0.95 |
| | Webtables | **0.84/0.95** | 0.50/0.80 | 0.69/0.92 | 0.70/**0.95** | 0.76/**0.95** |
| Nesting | CancerKG | **0.85/1.00** | 0.58/0.84 | 0.78/**1.00** | 0.60/0.90 | 0.70/0.88 |
| | CovidKG | **0.70/0.95** | 0.50/0.85 | 0.70/**0.95** | 0.50/0.90 | 0.60/0.86 |

Table 14: MAP/MRR for CC and TC with LLM – Textual and Numerical Content.

| Datasets | Tasks | Llama2 | GPT2 | RAG + Llama2 | RAG + GPT3.5 | RAG + GPT4 | TabBiN |
|---|---|---|---|---|---|---|---|
| String (CancerKG) | CC | 0.10/ 0.25 | 0.15/ 0.20 | 0.40/ **1.00** | 0.40/ **1.00** | 0.48/ **1.00** | **0.90/ 1.00** |
| | TC | 0.18/ 0.30 | 0.20/ 0.25 | 0.40/ **1.00** | 0.40/ 0.90 | 0.60/ **1.00** | **0.92/** 0.98 |
| Numerical (CovidKG) | CC | 0.12/ 0.20 | 0.10/ 0.20 | 0.30/ 0.90 | 0.25/ 0.90 | 0.35/ **1.00** | **0.60/** 0.90 |
| | TC | 0.17/ 0.20 | 0.20/ 0.20 | 0.30/ 0.90 | 0.30/ 0.90 | 0.38/ **1.00** | **0.53/** 0.90 |

From our experiments we conclude that RAG can be used both to improve LLM's performance on our downstream tasks as well as significantly reduce the size of the datasets processed by the LLM, which substantially reduces the cost of using commercial LLMs, especially for large-scale datasets. Alternative methods of more advanced prompting algorithms [72, 82] for complex tables could potentially enhance LLMs performance. This is one of the current directions of our further research.

## 5 RELATED WORK

The authors in [12] construct entity-centric embeddings for relational data. The embedding training sentence generation algorithm in [12] uses a graph, constructed per each *entity* found in tables (i.e., *Paul* in Figure 1). It does not take into account the intricacies of structure of the 2D neighboring context (i.e., *vertical* neighboring cells in the same column or *horizontal* in the same tuple) as well as does not distinguish *data* from *metadata*. [12] supports only *relational* tables, so it does not explicitly encode *hierarchical* metadata and does not distinguish between vertical metadata and data in non-relational tables. Similarly, it does not recognize nested tables or data values in different units, and treats numerical ranges as just 2 numbers, unlike us.

The authors train TABERT model [87] on Wikitables and show it outperforming BERT [22] on two benchmarks - SPIDER text-to-SQL [88] and WikiTableQuestions, "where a system has to infer latent DB queries from its execution results" [64]. Similarly, there are more questions answering models for tables [24, 33, 53, 84] built using a standard transformer architecture [22, 47] that use HybridQA [15], SQA, WikiSQL and WikiTQ [84] to evaluate standard questions answering tasks (QA) on data from semi-structured HTML and relational tables.

TabPrompt [37] adapts graph contrastive learning using Graph Neural Network (GNN) to encode tabular data and prompt-based learning to alleviate scarcity of labelled tabular training sets. Its performance is evaluated on two downstream tasks - cell and table type classification, similar to [80]. However, it does not support more complex non-relational tables, such as in this paper.

MotherNet [59] adapts the TabPFN [34] transformer architecture and focuses on supervised classification for small numeric tabular datasets from the OpenML-CC18 Benchmark [8]. It supports only *relational* tables and was not evaluated on any large-scale datasets with more complex tables as well as downstream tasks related to table structure understanding. Finally, it is *supervised*, which is a significant difference, since it requires labeled training data unlike us. However, there are studies focusing on generating labels for binary or multiclass classification of tabular datasets [2, 35, 51, 70, 81, 83, 85, 86, 91].

HYTEL [14] employs hypergraph-structure-aware transformer to encode tables and uses it for a series of downstream tasks, including column type annotation, column property annotation, table type detection and table similarity prediction (TSP). Authors utilized ≈ 1400 tables from PubMed Central (PMC) dataset [31] to evaluate TSP. StruBERT [75] also conducted table matching on the same dataset. However, unlike us, both methods fall short in providing exhaustive experimental evaluation on tables with multi-level hierarchical metadata and nesting. We also conducted all our evaluations on five large-scale datasets all from different domains.

TURL [21] is a (relational) structure-aware transformer, trained and evaluated on several tasks for table understanding,

such as relation extraction, row population, cell filling, schema augmentation, entity linking, and column type annotation. It also supports only *relational* tables, so it does not have a "special treatment" for hierarchical horizontal metadata as well as it treats vertical metadata as data. Similarly, it does not recognize nested tables or different units, and treats ranges as just 2 numbers.

Auto-Tables [48] learns a pipeline of data transformation operators using deep learning to transform non-relational tables into relational for query processing using SQL-based tools. Foofah [38], PATSQL [73], QBO [76], and Scythe [78] consider a subset of table-restructuring operators, which fall short in the Auto-Tables. In Auto-Tables, the authors work with non-relational tables, defined much more narrow than in this paper (Figure 1) and as we see them "in the wild". The non-relational tables in [48] lack hierarchy in metadata, nested tables, data values in different units for the same attribute, as well as numerical ranges. The transformation operators that the authors propose in [48] (*stack*, *wide-to-long*, *transpose*, *pivot*), therefore, are well-suited only for their simplified notion of non-relational tables.

[9] introduces an attribute-unionability framework that assesses table similarity by assessing their attribute relatedness. Aurum [27] leverages enterprise knowledge graph (EKG) to capture and query relationships among datasets in Data Lakes. It focuses on indexing and keyword-search to find related datasets in corporate data lakes based on simple matching of the terms from the users' query to the tables. Our semantic matching works based on the cosine similarity of composite embedding vectors for non-relational tables that incorporate all components of such tables separately – hierarchical metadata and data, nested tables, inferred types, units of data values, ranges, etc. Such complex vectors are composed in order to preserve semantic differences of each component. This, in turn, affects quality of matching with and without such vectors.

Tabllm [32] fine-tuned T0 [6] and GPT-3 [62] models for tabular classification. These LLMs demonstrated competitive performance, comparable to baselines, such as gradient-boosted trees, on OpenML tabular datasets [8]. [72] introduces a benchmark that evaluates LLMs (GPT-3.5 [72] and GPT-4 [54]) on seven tabular downstream tasks, such as column retrieval and cell lookup, utilizing various LLM prompt designs and table input formatting. TapTap [90] uses GPT-2 [66] to encode single rows independently using a "text template serialization" strategy, resulting in singular row embeddings. They can be used in several downstream tasks, such as table data augmentation, imputation, and handling imbalanced classification. All these studies [32, 68, 72, 89] focus on relational tables, unlike ours. However, the authors formulate interesting insights on capabilities [32, 89] and limitations [68, 72] of current LLMs in table understanding. In [72], p.2. the authors state *"LLMs have basic structural understanding capabilities but are far from perfect, even on trivial tasks, e.g., table size detection (detect the number of columns and rows in a table)"*. By carefully choosing the LLM input (e.g., table input format, content order, role prompting, and partition marks) and different prompt designs, the authors achieved promising improvements in structural understanding capabilities of LLMs. In [68], the authors investigated inconsistencies in GPT3 performance in self-supervised structural table understanding tasks (e.g., table transposition, column reordering) depending on the data format (i.e., HTML, JSON, CSV, DFLoader, etc.) and noise-operations (e.g., merging cells, shuffling column names). They demonstrate new possibilities of using LLMs for structured data understanding via effective prompt design.

NumSearchLLM [71] also leverages LLMs (GPT-3.5 and Llama2 [74]) as well as enterprise Knowledge Graphs to perform table search over purely *numeric* tables. [82] proposes Chain-of-Table method for table understanding tasks, such as table-based question answering and fact verification. It dynamically updates the table content in the reasoning process by employing LLMs to iteratively generate SQL-like table operations such as adding columns, selecting rows, grouping, and more. The resulting table is then fed back to the LLMs to generate the final answer. In contrast, our method focuses on learning the fine-grained embedding representation, optimized for *non-relational* tables having hierarchical HMD, VMD, and nesting and using them to perform high accuracy table, column, and entity clustering/matching.

[25] extends data discovery process in Data Lakes across two modalities of structured and unstructured data using a model capturing similarities between text documents $d$ and tabular columns $c$. To train such a model, the authors curate a labeled training set indicating the relation between $d$ and $c$. Their application spans from document-to-table relatedness to table-to-table relatedness. We evaluate our embeddings on downstream tasks including column-to-column and table-to-table similarity. Our embeddings have fine-grained structure taking into account the finest intricacies of non-relational tables as discussed in the previous paragraph. Our approach is also unsupervised, hence does not need labelling.

Tabbie [36] and TUTA [80] train embeddings and evaluate them on several different downstream tasks — row population, column population, column type prediction, cell and table type classification. Unlike Tabbie, TUTA and other SOTA solutions for relational tables, TabBiN supports *complex non-relational tables* with nesting, distinguishes data and metadata context, recognizes both vertical and horizontal hierarchical metadata, performs type inference on both metadata and data, uniquely embeds not only numerical values but also ranges, recognizes *units* and encodes them as separate embeddings vectors.

## 6 CONCLUSION

We introduced TabBiN – a structure- and metadata-aware transformer for tables not in 1st Normal Form with hierarchical vertical and horizontal metadata, having nested tables, data values in different units, and numerical ranges. We refer to them as non-relational or BiN tables. Relational tables constitute only 0.9% of all tables in the common Web crawl [46] and 22% of spreadsheet tables, while the rest are non-relational. To the best of our knowledge, TabBiN is the first transformer-based unsupervised architecture optimized for intricacies of structural context in these tables, respecting units in numerical values, and treating ranges and gaussians according to their semantics, not blindly as a sequence of numbers as in many SOTA solutions. TabBiN also performs semantic type inference on the table content as well as its metadata and encodes inferred types as an additional component in the embedding layer. This fine-grained understanding and "special treatment" of non-relational tables with hierarchical metadata and nesting, allows TabBiN to outperform SOTA on three popular downstream tasks on five large-scale structured datasets with the significant MAP delta of up to 0.28. GPT-4 LLM+RAG slightly outperforms us with MRR delta of 0.1, but we significantly outperformed it with the MAP delta of up to 0.42.

# ACKNOWLEDGMENTS

We would like to thank Amazon and NSF (Award # 2345794) for supporting fundamental parts of research on this project, Casey DeSantis Cancer Innovation Fund, Florida Department of Health (DOH) for their support of both Computer Science/AI research as well as its application to Cancer. We thank anonymous reviewers for their valuable feedback.

# REFERENCES

[1] 2024. Aryn-Ai/Sycamore: Sycamore is an LLM-Powered Search and Analytics Platform for Unstructured Data. https://github.com/aryn-ai/sycamore.

[2] Md Atik Ahamed and Qiang Cheng. 2024. Mambatab: A simple yet effective approach for handling tabular data. *arXiv preprint arXiv:2401.08867* (2024).

[3] Bogdan Alexe, Michael Gubanov, Mauricio A Hernández, Howard Ho, Jen-Wei Huang, Yannis Katsis, Lucian Popa, Barna Saha, and Ioana Stanoi. 2009. Simplifying information integration: Object-based flow-of-mappings framework for integration. In *Business Intelligence for the Real-Time Enterprise: Second International Workshop, BIRTE 2008, Auckland, New Zealand, August 24, 2008, Revised Selected Papers 2*. Springer, 108–121.

[4] Sercan Ö Arik and Tomas Pfister. 2021. Tabnet: Attentive interpretable tabular learning. In *Proceedings of the AAAI conference on artificial intelligence*, Vol. 35. 6679–6687.

[5] Vaswani Ashish. 2017. Attention is all you need. *Advances in neural information processing systems* 30 (2017), I.

[6] Stephen Bach, Victor Sanh, Zheng Xin Yong, Albert Webson, Colin Raffel, Nihal V Nayak, Abheesht Sharma, Taewoon Kim, M Saiful Bari, Thibault Fevry, et al. 2022. PromptSource: An Integrated Development Environment and Repository for Natural Language Prompts. In *Proceedings of the 60th Annual Meeting of the Association for Computational Linguistics: System Demonstrations*. 93–104.

[7] Chandra Sekhar Bhagavatula, Thanapon Noraset, and Doug Downey. 2015. Tabel: Entity linking in web tables. In *International Semantic Web Conference*. Springer, 425–441.

[8] Bernd Bischl, Giuseppe Casalicchio, Matthias Feurer, Pieter Gijsbers, Frank Hutter, Michel Lang, Rafael G Mantovani, Jan N van Rijn, and Joaquin Vanschoren. 2017. Openml benchmarking suites. *arXiv preprint arXiv:1708.03731* (2017).

[9] Alex Bogatu, Alvaro AA Fernandes, Norman W Paton, and Nikolaos Konstantinou. 2020. Dataset discovery in data lakes. In *2020 ieee 36th international conference on data engineering (icde)*. IEEE, 709–720.

[10] Michael J Cafarella. [n.d.]. Uncovering the Relational Web. Citeseer.

[11] Michael J Cafarella, Alon Halevy, Daisy Zhe Wang, Eugene Wu, and Yang Zhang. 2008. Webtables: exploring the power of tables on the web. *Proceedings of the VLDB Endowment* 1, 1 (2008), 538–549.

[12] Riccardo Cappuzzo, Paolo Papotti, and Saravanan Thirumuruganathan. 2020. Creating embeddings of heterogeneous relational datasets for data integration tasks. In *Proceedings of the 2020 ACM SIGMOD international conference on management of data*. 1335–1349.

[13] Maitry Chauhan, Anna Pyayt, and Michael Gubanov. 2023. Learning Topical Structured Interfaces from Medical Research Literature. In *Companion Proceedings of the ACM Web Conference 2023*. 225–228.

[14] Pei Chen, Soumajyoti Sarkar, Leonard Lausen, Balasubramaniam Srinivasan, Sheng Zha, Ruihong Huang, and George Karypis. 2024. HYTREL: Hypergraph-enhanced tabular data representation learning. *Advances in Neural Information Processing Systems* 36 (2024).

[15] Wenhu Chen, Hanwen Zha, Zhiyu Chen, Wenhan Xiong, Hong Wang, and William Yang Wang. 2020. HybridQA: A Dataset of Multi-Hop Question Answering over Tabular and Textual Data. In *Findings of the Association for Computational Linguistics: EMNLP 2020*. 1026–1036.

[16] Zhe Chen and Michael Cafarella. 2014. Integrating spreadsheet data via accurate and low-effort extraction. In *Proceedings of the 20th ACM SIGKDD international conference on Knowledge discovery and data mining*. 1126–1135.

[17] Hyung Won Chung, Le Hou, Shayne Longpre, Barret Zoph, Yi Tay, William Fedus, Yunxuan Li, Xuezhi Wang, Mostafa Dehghani, Siddhartha Brahma, et al. 2024. Scaling instruction-finetuned language models. *Journal of Machine Learning Research* 25, 70 (2024), 1–53.

[18] Edgar F Codd. 1972. Further normalization of the data base relational model. *Data base systems* 6 (1972), 33–64.

[19] Tianji Cong, Fatemeh Nargesian, and HV Jagadish. 2023. Pylon: Semantic Table Union Search in Data Lakes. *arXiv preprint arXiv:2301.04901* (2023).

[20] Nick Craswell. 2009. Mean reciprocal rank. *Encyclopedia of database systems* (2009), 1703–1703.

[21] Xiang Deng, Huan Sun, Alyssa Lees, You Wu, and Cong Yu. 2022. Turl: Table understanding through representation learning. *ACM SIGMOD Record* 51, 1 (2022), 33–40.

[22] Jacob Devlin, Ming-Wei Chang, Kenton Lee, and Kristina Toutanova. 2018. Bert: Pre-training of deep bidirectional transformers for language understanding. *arXiv preprint arXiv:1810.04805* (2018).

[23] Haoyu Dong, Shijie Liu, Zhouyu Fu, Shi Han, and Dongmei Zhang. 2019. Semantic structure extraction for spreadsheet tables with a multi-task learning architecture. In *Workshop on Document Intelligence at NeurIPS 2019*.

[24] Julian Eisenschlos, Maharshi Gor, Thomas Mueller, and William Cohen. 2021. MATE: Multi-view Attention for Table Transformer Efficiency. In *Proceedings of the 2021 Conference on Empirical Methods in Natural Language Processing*. 7606–7619.

[25] Mohamed Y Eltabakh, Mayuresh Kunjir, Ahmed K Elmagarmid, and Mohammad Shahmeer Ahmad. 2023. Cross Modal Data Discovery over Structured and Unstructured Data Lakes. *Proceedings of the VLDB Endowment* 16, 11 (2023), 3377–3390.

[26] Alessandro Fantechi, Stefania Gnesi, Samuele Livi, and Laura Semini. 2021. A spaCy-based tool for extracting variability from NL requirements. In *Proceedings of the 25th ACM International Systems and Software Product Line Conference-Volume B*. 32–35.

[27] Raul Castro Fernandez, Ziawasch Abedjan, Famien Koko, Gina Yuan, Samuel Madden, and Michael Stonebraker. 2018. Aurum: A data discovery system. In *2018 IEEE 34th International Conference on Data Engineering (ICDE)*. IEEE, 1001–1012.

[28] Anna Lisa Gentile, Petar Ristoski, Steffen Eckel, Dominique Ritze, and Heiko Paulheim. 2017. Entity matching on web tables: a table embeddings approach for blocking. (2017).

[29] Majid Ghasemi-Gol and Pedro Szekely. 2018. Tabvec: Table vectors for classification of web tables. *arXiv preprint arXiv:1802.06290* (2018).

[30] Majid Ghasemi Gol, Jay Pujara, and Pedro Szekely. 2019. Tabular cell classification using pre-trained cell embeddings. In *2019 IEEE International Conference on Data Mining (ICDM)*. IEEE, 230–239.

[31] Maryam Habibi, Johannes Starlinger, and Ulf Leser. 2020. Tabsim: A siamese neural network for accurate estimation of table similarity. In *2020 ieee international conference on big data (big data)*. IEEE, 930–937.

[32] Stefan Hegselmann, Alejandro Buendia, Hunter Lang, Monica Agrawal, Xiaoyi Jiang, and David Sontag. 2023. Tabllm: Few-shot classification of tabular data with large language models. In *International Conference on Artificial Intelligence and Statistics*. PMLR, 5549–5581.

[33] Jonathan Herzig, Pawel Krzysztof Nowak, Thomas Mueller, Francesco Piccinno, and Julian Eisenschlos. 2020. TaPas: Weakly Supervised Table Parsing via Pre-training. In *Proceedings of the 58th Annual Meeting of the Association for Computational Linguistics*. 4320–4333.

[34] Noah Hollmann, Samuel Müller, Katharina Eggensperger, and Frank Hutter. 2022. Tabpfn: A transformer that solves small tabular classification problems in a second. *arXiv preprint arXiv:2207.01848* (2022).

[35] Xin Huang, Ashish Khetan, Milan Cvitkovic, and Zohar Karnin. 2020. Tabtransformer: Tabular data modeling using contextual embeddings. *arXiv preprint arXiv:2012.06678* (2020).

[36] Hiroshi Iida, Dung Thai, Varun Manjunatha, and Mohit Iyyer. 2021. TABBIE: Pretrained Representations of Tabular Data. In *Proceedings of the 2021 Conference of the North American Chapter of the Association for Computational Linguistics: Human Language Technologies*.

[37] Rihui Jin, Jianan Wang, Wei Tan, Yongrui Chen, Guilin Qi, and Wang Hao. 2023. TabPrompt: Graph-based Pre-training and Prompting for Few-shot Table Understanding. In *Findings of the Association for Computational Linguistics: EMNLP 2023*. 7373–7383.

[38] Zhongjun Jin, Michael R Anderson, Michael Cafarella, and HV Jagadish. 2017. Foofah: Transforming data by example. In *Proceedings of the 2017 ACM International Conference on Management of Data*. 683–698.

[39] Jacqueline Joseph. 2023. Assessing the potential of laboratory instructional tool through Synthesia AI: a case study on student learning outcome. *International Journal of e-Learning and Higher Education (IJELHE)* 18, 2 (2023), 5–16.

[40] Bhimesh Kandibedala, Anna Pyayt, Chris Caballero, and Michael N Gubanov. 2023. Scalable Hierarchical Metadata Classification in Heterogeneous Large-scale Datasets.. In *DOLAP*. 81–85.

[41] Bhim Kandibedala, Anna Pyayt, Nick Piraino, Chris Caballero, and Michael Gubanov. 2023. COVIDKG. ORG-a Web-scale COVID-19 Interactive, Trustworthy Knowledge Graph, Constructed and Interrogated for Bias using Deep-Learning. In *International Conference on Extending Database Technology (EDBT)*.

[42] Pradap Konda, Sanjib Das, Paul Suganthan GC, AnHai Doan, Adel Ardalan, Jeffrey R Ballard, Han Li, Fatemah Panahi, Haojun Zhang, Jeff Naughton, et al. 2016. Magellan: Toward Building Entity Matching Management Systems. *Proceedings of the VLDB Endowment* 9, 12 (2016).

[43] Hanna Köpcke, Andreas Thor, and Erhard Rahm. 2010. Evaluation of entity resolution approaches on real-world match problems. *Proceedings of the VLDB Endowment* 3, 1-2 (2010), 484–493.

[44] Guillaume Lample and Alexis Conneau. 2019. Cross-lingual language model pretraining. *arXiv preprint arXiv:1901.07291* (2019).

[45] Jinhyuk Lee, Wonjin Yoon, Sungdong Kim, Donghyeon Kim, Sunkyu Kim, Chan Ho So, and Jaewoo Kang. 2020. BioBERT: a pre-trained biomedical language representation model for biomedical text mining. *Bioinformatics* 36, 4 (2020), 1234–1240.

[46] Oliver Lehmberg, Dominique Ritze, Robert Meusel, and Christian Bizer. 2016. A large public corpus of web tables containing time and context metadata. In *Proceedings of the 25th international conference companion on world wide web*. 75–76.

[47] Mike Lewis, Yinhan Liu, Naman Goyal, Marjan Ghazvininejad, Abdelrahman Mohamed, Omer Levy, Veselin Stoyanov, and Luke Zettlemoyer. 2020. BART: Denoising Sequence-to-Sequence Pre-training for Natural Language Generation, Translation, and Comprehension. In *Proceedings of the 58th Annual*

*Meeting of the Association for Computational Linguistics*. 7871–7880.

[48] Peng Li, Yeye He, Cong Yan, Yue Wang, and Surajit Chaudhuri. 2023. Auto-Tables: Synthesizing Multi-Step Transformations to Relationalize Tables without Using Examples. *Proceedings of the VLDB Endowment* 16, 11 (2023), 3391–3403.

[49] Yuliang Li, Jinfeng Li, Yoshihiko Suhara, AnHai Doan, and Wang-Chiew Tan. 2020. Deep entity matching with pre-trained language models. *Proceedings of the VLDB Endowment* 14, 1 (2020), 50–60.

[50] Seung-Jin Lim and Yiu-Kai Ng. 1999. An automated approach for retrieving hierarchical data from HTML tables. In *Proceedings of the eighth international conference on Information and knowledge management*. 466–474.

[51] Guang Liu, Jie Yang, and Ledell Wu. 2022. Ptab: Using the pre-trained language model for modeling tabular data. *arXiv preprint arXiv:2209.08060* (2022).

[52] Ling Liu, M Tamer Özsu, et al. 2009. Mean average precision. *Encyclopedia of database systems* 1703 (2009).

[53] Qian Liu, Bei Chen, Jiaqi Guo, Morteza Ziyadi, Zeqi Lin, Weizhu Chen, and Jian-Guang Lou. 2021. TAPEX: Table pre-training via learning a neural SQL executor. *arXiv preprint arXiv:2107.07653* (2021).

[54] Yiheng Liu, Tianle Han, Siyuan Ma, Jiayue Zhang, Yuanyuan Yang, Jiaming Tian, Hao He, Antong Li, Mengshen He, Zhengliang Liu, et al. 2023. Summary of chatgpt-related research and perspective towards the future of large language models. *Meta-Radiology* (2023), 100017.

[55] Yinhan Liu, Myle Ott, Naman Goyal, Jingfei Du, Mandar Joshi, Danqi Chen, Omer Levy, Mike Lewis, Luke Zettlemoyer, and Veselin Stoyanov. 2019. Roberta: A robustly optimized bert pretraining approach. *arXiv preprint arXiv:1907.11692* (2019).

[56] L Lu Wang, Kyle Lo, Yoganand Chandrasekhar, Russell Reas, Jiangjiang Yang, Darrin Eide, Kathryn Funk, Rodney Kinney, Ziyang Liu, William Merrill, et al. 2020. CORD-19: The Covid-19 open research dataset. ArXiv. *arXiv preprint arXiv:2004.10706* (2020).

[57] Sergey Melnik, Erhard Rahm, and Philip A Bernstein. 2003. Rondo: A programming platform for generic model management. In *Proceedings of the 2003 ACM SIGMOD international conference on Management of data*. 193–204.

[58] Tomas Mikolov, Ilya Sutskever, Kai Chen, Greg S Corrado, and Jeff Dean. 2013. Distributed representations of words and phrases and their compositionality. *Advances in neural information processing systems* 26 (2013).

[59] Andreas Müller, Carlo Curino, and Raghu Ramakrishnan. 2023. MotherNet: A Foundational Hypernetwork for Tabular Classification. *arXiv preprint arXiv:2312.08598* (2023).

[60] Mark Neumann, Daniel King, Iz Beltagy, and Waleed Ammar. 2019. ScispaCy: Fast and Robust Models for Biomedical Natural Language Processing. *BioNLP 2019* (2019), 319.

[61] Xuan-Phi Nguyen, Shafiq Joty, Steven CH Hoi, and Richard Socher. 2020. Tree-structured attention with hierarchical accumulation. *arXiv preprint arXiv:2002.08046* (2020).

[62] Long Ouyang, Jeffrey Wu, Xu Jiang, Diogo Almeida, Carroll Wainwright, Pamela Mishkin, Chong Zhang, Sandhini Agarwal, Katarina Slama, Alex Ray, et al. 2022. Training language models to follow instructions with human feedback. *Advances in neural information processing systems* 35 (2022), 27730–27744.

[63] Viacheslav Paramonov, Alexey Shigarov, and Varvara Vetrova. 2020. Table header correction algorithm based on heuristics for improving spreadsheet data extraction. In *International Conference on Information and Software Technologies*. Springer, 147–158.

[64] Panupong Pasupat and Percy Liang. 2015. Compositional Semantic Parsing on Semi-Structured Tables. In *Proceedings of the 53rd Annual Meeting of the Association for Computational Linguistics and the 7th International Joint Conference on Natural Language Processing (Volume 1: Long Papers)*. 1470–1480.

[65] Sophie Pavia, Montasir Shams, Rituparna Khan, Anna Pyayt, and Michael Gubanov. 2021. Learning Tabular Embeddings at Web Scale. In *2021 IEEE International Conference on Big Data (Big Data)*. IEEE, 4936–4944.

[66] Alec Radford, Jeffrey Wu, Rewon Child, David Luan, Dario Amodei, Ilya Sutskever, et al. 2019. Language models are unsupervised multitask learners. *OpenAI blog* 1, 8 (2019), 9.

[67] Vighnesh Shiv and Chris Quirk. 2019. Novel positional encodings to enable tree-based transformers. *Advances in neural information processing systems* 32 (2019).

[68] Ananya Singha, José Cambronero, Sumit Gulwani, Vu Le, and Chris Parnin. [n.d.]. Tabular Representation, Noisy Operators, and Impacts on Table Structure Understanding Tasks in LLMs. In *NeurIPS 2023 Second Table Representation Learning Workshop*.

[69] Sean Soderman, Anusha Kola, Maksim Podkorytov, Michael Geyer, and Michael Gubanov. 2018. Hybrid. ai: A learning search engine for large-scale structured data. In *Companion Proceedings of the The Web Conference 2018*. 1507–1514.

[70] Gowthami Somepalli, Avi Schwarzschild, Micah Goldblum, C Bayan Bruss, and Tom Goldstein. [n.d.]. SAINT: Improved Neural Networks for Tabular Data via Row Attention and Contrastive Pre-Training. In *NeurIPS 2022 First Table Representation Workshop*.

[71] Pranav Subramaniam, Udayan Khurana, Kavitha Srinivas, and Horst Samulowitz. 2023. Related Table Search for Numeric data using Large Language Models and Enterprise Knowledge Graphs. In *ACM International Conference on Information and Knowledge Management*.

[72] Yuan Sui, Mengyu Zhou, Mingjie Zhou, Shi Han, and Dongmei Zhang. 2024. Table meets llm: Can large language models understand structured table data? a benchmark and empirical study. In *Proceedings of the 17th ACM International Conference on Web Search and Data Mining*. 645–654.

[73] Keita Takenouchi, Takashi Ishio, Joji Okada, and Yuji Sakata. 2021. PATSQL: efficient synthesis of SQL queries from example tables with quick inference of projected columns. *Proceedings of the VLDB Endowment* 14, 11 (2021), 1937–1949.

[74] Hugo Touvron, Louis Martin, Kevin Stone, Peter Albert, Amjad Almahairi, Yasmine Babaei, Nikolay Bashlykov, Soumya Batra, Prajjwal Bhargava, Shruti Bhosale, et al. 2023. Llama 2: Open foundation and fine-tuned chat models. *arXiv preprint arXiv:2307.09288* (2023).

[75] Mohamed Trabelsi, Zhiyu Chen, Shuo Zhang, Brian D Davison, and Jeff Heflin. 2022. Strubert: Structure-aware bert for table search and matching. In *Proceedings of the ACM Web Conference 2022*. 442–451.

[76] Quoc Trung Tran, Chee-Yong Chan, and Srinivasan Parthasarathy. 2009. Query by output. In *Proceedings of the 2009 ACM SIGMOD International Conference on Management of data*. 535–548.

[77] Helena Verdaguer, Josep Tabernero, and Teresa Macarulla. 2016. Ramucirumab in metastatic colorectal cancer: evidence to date and place in therapy. *Therapeutic advances in medical oncology* 8, 3 (2016), 230–242.

[78] Chenglong Wang, Alvin Cheung, and Rastislav Bodik. 2017. Synthesizing highly expressive SQL queries from input-output examples. In *Proceedings of the 38th ACM SIGPLAN Conference on Programming Language Design and Implementation*. 452–466.

[79] Yaushian Wang, Hung-Yi Lee, and Yun-Nung Chen. 2019. Tree Transformer: Integrating Tree Structures into Self-Attention. In *Proceedings of the 2019 Conference on Empirical Methods in Natural Language Processing and the 9th International Joint Conference on Natural Language Processing (EMNLP-IJCNLP)*. 1061–1070.

[80] Zhiruo Wang, Haoyu Dong, Ran Jia, Jia Li, Zhiyi Fu, Shi Han, and Dongmei Zhang. 2021. Tuta: Tree-based transformers for generally structured table pre-training. In *Proceedings of the 27th ACM SIGKDD Conference on Knowledge Discovery & Data Mining*. 1780–1790.

[81] Zifeng Wang and Jimeng Sun. 2022. Transtab: Learning transferable tabular transformers across tables. *Advances in Neural Information Processing Systems* 35 (2022), 2902–2915.

[82] Zilong Wang, Hao Zhang, Chun-Liang Li, Julian Martin Eisenschlos, Vincent Perot, Zifeng Wang, Lesly Miculicich, Yasuhisa Fujii, Jingbo Shang, Chen-Yu Lee, et al. 2024. Chain-of-Table: Evolving Tables in the Reasoning Chain for Table Understanding. In *The Twelfth International Conference on Learning Representations*.

[83] Witold Wydmański, Oleksii Bulenok, and Marek Śmieja. 2023. Hypertab: Hypernetwork approach for deep learning on small tabular datasets. In *2023 IEEE 10th International Conference on Data Science and Advanced Analytics (DSAA)*. IEEE, 1–9.

[84] Jingfeng Yang, Aditya Gupta, Shyam Upadhyay, Luheng He, Rahul Goel, and Shachi Paul. 2022. TableFormer: Robust Transformer Modeling for Table-Text Encoding. In *Proceedings of the 60th Annual Meeting of the Association for Computational Linguistics (Volume 1: Long Papers)*. 528–537.

[85] Yazheng Yang, Yuqi Wang, Guang Liu, Ledell Wu, and Qi Liu. 2023. Unitabe: Pretraining a unified tabular encoder for heterogeneous tabular data. *arXiv preprint arXiv:2307.09249* (2023).

[86] Chao Ye, Guoshan Lu, Haobo Wang, Liyao Li, Sai Wu, Gang Chen, and Junbo Zhao. 2023. CT-BERT: learning better tabular representations through cross-table pre-training. *arXiv preprint arXiv:2307.04308* (2023).

[87] Pengcheng Yin, Graham Neubig, Wen-tau Yih, and Sebastian Riedel. 2020. TaBERT: Pretraining for joint understanding of textual and tabular data. *arXiv preprint arXiv:2005.08314* (2020).

[88] Tao Yu, Rui Zhang, Kai Yang, Michihiro Yasunaga, Dongxu Wang, Zifan Li, James Ma, Irene Li, Qingning Yao, Shanelle Roman, et al. 2018. Spider: A Large-Scale Human-Labeled Dataset for Complex and Cross-Domain Semantic Parsing and Text-to-SQL Task. In *Proceedings of the 2018 Conference on Empirical Methods in Natural Language Processing*. 3911–3921.

[89] Liangyu Zha, Junlin Zhou, Liyao Li, Rui Wang, Qingyi Huang, Saisai Yang, Jing Yuan, Changbao Su, Xiang Li, Aofeng Su, et al. 2023. Tablegpt: Towards unifying tables, nature language and commands into one gpt. *arXiv preprint arXiv:2307.08674* (2023).

[90] Tianping Zhang, Shaowen Wang, Shuicheng Yan, Li Jian, and Qian Liu. 2023. Generative Table Pre-training Empowers Models for Tabular Prediction. In *Proceedings of the 2023 Conference on Empirical Methods in Natural Language Processing*. 14836–14854.

[91] Bingzhao Zhu, Xingjian Shi, Nick Erickson, Mu Li, George Karypis, and Mahsa Shoaran. 2023. XTab: cross-table pretraining for tabular transformers. In *Proceedings of the 40th International Conference on Machine Learning*. 43181–43204.